\documentclass[letterpaper]{article} 
\usepackage{aaai25}  
\usepackage{times}  
\usepackage{helvet}  
\usepackage{courier}  
\usepackage[hyphens]{url}  
\usepackage{graphicx} 
\usepackage{natbib}  
\usepackage{caption} 
\usepackage{algorithm}
\usepackage{algorithmic}
\usepackage{amsmath,amssymb,amsfonts}
\usepackage{algorithmic}
\usepackage{textcomp}
\usepackage{xcolor}
\usepackage{booktabs}
\usepackage{graphicx,color}
\usepackage{multirow}
\usepackage{bm} 
\usepackage{bbding}
\usepackage{newfloat}
\usepackage{listings}
\DeclareCaptionStyle{ruled}{labelfont=normalfont,labelsep=colon,strut=off} 
\floatstyle{ruled}
\newfloat{listing}{tb}{lst}{}
\floatname{listing}{Listing}
\title{Qsco: A Quantum Scoring Module for Open-set Supervised Anomaly Detection}
\author{
    Yifeng Peng\textsuperscript{\rm1}, Xinyi Li\textsuperscript{\rm1}, Zhiding Liang\textsuperscript{\rm2}, Ying Wang\textsuperscript{\rm1}\thanks{Corresponding Author.}
}
\affiliations{
    \textsuperscript{\rm 1}Stevens Institute of Technology, Hoboken, NJ, USA\\
    \textsuperscript{\rm2}Rensselaer Polytechnic Institute, Troy, NY, USA \\
    \{ypeng21, xli215, ywang6\}@stevens.edu, liangz9@rpi.edu

}

\usepackage{bibentry}

\begin{document}

\maketitle

\begin{abstract}
Open set anomaly detection (OSAD) is a crucial task that aims to identify abnormal patterns or behaviors in data sets, especially when the anomalies observed during training do not represent all possible classes of anomalies. The recent advances in quantum computing in handling complex data structures and improving machine learning models herald a paradigm shift in anomaly detection methodologies. This study proposes a Quantum Scoring Module (Qsco), embedding quantum variational circuits into neural networks to enhance the model's processing capabilities in handling uncertainty and unlabeled data. Extensive experiments conducted across eight real-world anomaly detection datasets demonstrate our model's superior performance in detecting anomalies across varied settings and reveal that integrating quantum simulators does not result in prohibitive time complexities. At the same time, the experimental results under different noise models also prove that Qsco is a noise-resilient algorithm. Our study validates the feasibility of quantum-enhanced anomaly detection methods in practical applications. 
\end{abstract}

\section{Introduction}

Precisely identifying anomalous patterns within extensive and intricate datasets is paramount, especially when data labels are nonexistent, or the underlying data distributions remain undefined. Such conditions delineate the open set anomaly detection (OSAD) field, characterized by its necessity to function robustly across varied and often unforeseen data landscapes. Historically, the challenges inherent in OSAD have been addressed through conventional machine learning strategies \cite{ding2022catching, SAOE1, SAOE2,pang2021explainable,pang2019deep}. While these methods have proven effective, they increasingly encounter constraints related to data's escalating scale and complexity.

\begin{figure}[ht]

\begin{center}
\centering
\centerline{\includegraphics[width=0.5\textwidth]{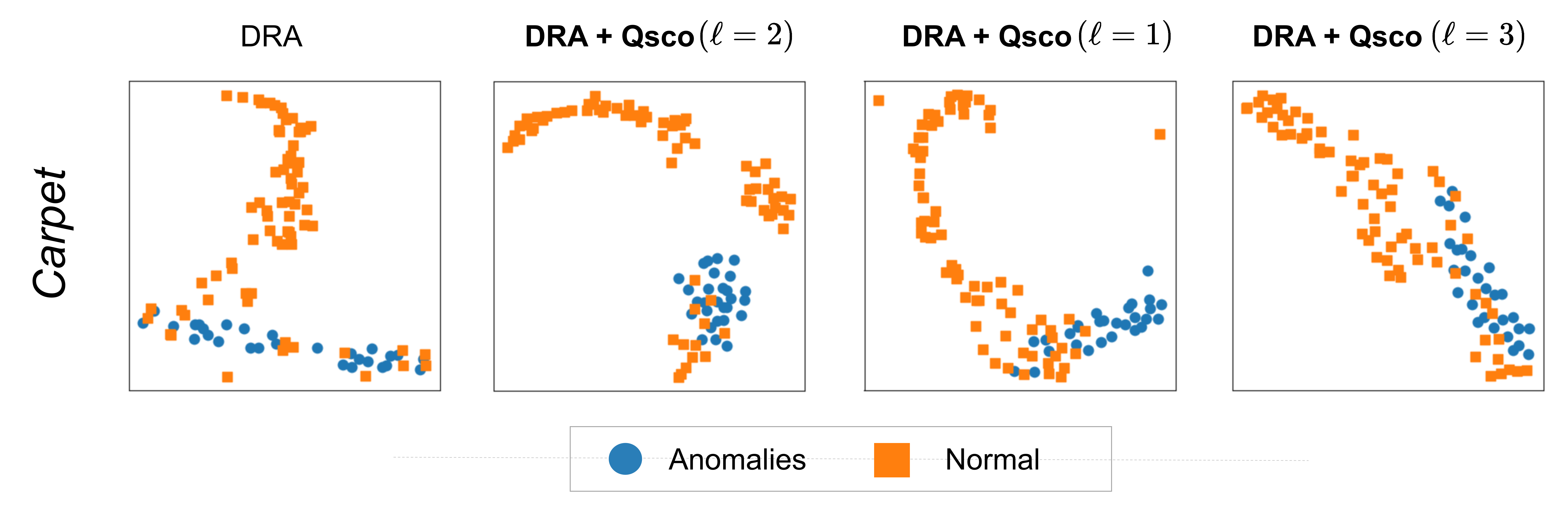}}
\caption{Visualization of the anomalies and normal data predicted by DRA \cite{ding2022catching} and \textbf{Qsco (Ours)} in the MVTec AD \cite{Bergmann2019MVTecA} dataset (carpet subset). The parameter $\ell$ controls the depth of the variational circuit in Qsco. With the correct $\ell$ level, Qsco enhances DRA's ability to distinguish boundaries between anomalies and normal data. However, if $\ell$ is too high ($\ell = 3$), it results in over-fitting, as shown in the last graph, while if $\ell$ is too low ($\ell = 1$), it leads to under-fitting.}
\label{distribution}
\end{center}
\end{figure}

The advent of quantum computing has foreboded new prospects for augmenting machine learning architectures \cite{pan2023hybrid, caro2022generalization}. Quantum computing offers substantial enhancements in computational efficiency and capacity, particularly adept at managing the high-dimensional data spaces prevalent in anomaly detection tasks. The synergy of quantum computing with machine learning, commonly referred to as quantum machine learning (QML) \cite{Farhi2018ClassificationWQqnnqml,alchieri2021introduction,peng2024hyq2}, introduces innovative approaches that may surmount some of the fundamental limitations of classical methodologies, including scalability, complexity, and uncertainty. With advancements in quantum hardware, QML is poised for revolutionary applications in AI domains like quantum natural language processing (QNLP) \cite{abbaszade2021application}, quantum computer vision (QCV) \cite{wei2023quantum}, and quantum anomaly detection (QAD)\cite{liu2018quantum}.

This research introduces the Quantum Scoring Module (Qsco), a state-of-the-art quantum model of quantum variational circuits with different rotation and CNOT gates. This model aims to exploit the quantum-enhanced computational power to refine the detection of obscure patterns and anomalies in open-set contexts. With the help of entanglement characteristics in quantum computing, we can discover the distribution of unknown anomalies and solve distribution problems that classical computing cannot simulate. We provide initial findings that indicate the Qsco model's enhanced efficacy in processing high-dimensional data relative to traditional models.

In this paper, we investigate the application of QML in OSAD and endeavor to serve as a foundational text for subsequent inquiries into the fusion of quantum and classical computing paradigms. Through this scholarly endeavor, we aspire to forge the development of novel anomaly detection frameworks that are more competent, precise, and equipped to handle the multifaceted demands of contemporary data landscapes.

\section{Related Work}
\textbf{Open Set Anomaly Detection} The recent advancements in open set recognition have led to various innovative approaches to detect and classify unknown samples effectively. \cite{MLEP} introduced the Margin Learning Embedded Prediction (MLEP) framework, emphasizing the importance of margin learning for enhancing open set recognition. Building on this, \cite{acsintoae2022ubnormal} proposed a general open-set method, GOAD, which uses random affine transformations to extend the applicability of transformation-based methods to non-image data, thus relaxing current generalization assumptions. \cite{sun2020conditional} tackled the limitations of variational auto-encoders (VAEs) in providing discriminative representations for known classifications by proposing Conditional Gaussian Distribution Learning (CGDL), which enhances open set recognition. \cite{kong2021opengan} demonstrated that a well-chosen GAN discriminator on real outlier data can achieve state-of-the-art results and further improved performance by augmenting real open training examples with adversarially synthesized "fake" data, culminating in the OpenGAN framework. \cite{yoshihashi2019classification}) focused on joint classification and reconstruction of input data through their Classification-Reconstruction learning for Open-Set Recognition (CROSR), which utilizes latent representations to enhance the separation of unknowns from knowns without compromising known-class classification accuracy. Lastly, \cite{zhang2020hybrid} proposed the OpenHybrid framework, which combines an encoder, a classifier, and a flow-based density estimator to detect unknown samples while classifying inliers accurately and effectively.

\textbf{Supervised Anomaly Detection} Recent anomaly detection advancements have introduced various innovative methods and frameworks to improve how anomalies are identified and classified across various domains. \cite{gornitz2013toward} established a foundation with their supervised approach, paving the way for further research such as \cite{ruff2019deep}, who enhanced accuracy through deep semi-supervised methods leveraging both labeled and unlabeled data. \cite{li2021cutpaste} introduced Cutpaste, a self-supervised technique that emphasized the growing relevance of self-supervision in anomaly detection.

Further innovations include \cite{akcay2019ganomaly}, who developed Ganomaly using GANs for semi-supervised anomaly detection, and \cite{han2022adbench}, who created Adbench, a comprehensive benchmark for method evaluation. In video anomaly detection, \cite{georgescu2021anomaly} and \cite{tian2021weakly} demonstrated the effectiveness of self-supervised and weakly-supervised learning for handling complex data scenarios and robust temporal feature extraction, respectively.

Additionally, \cite{tang2024gadbench} and \cite{Yao_2023_CVPR} contributed by benchmarking graph-based methods with Gadbench and enhancing supervised detection with boundary-guided contrastive learning. In the realm of weak supervision, \cite{cho2023look} and \cite{bozorgtabar2023attention} explored relational and attention-conditioned learning approaches, while \cite{chen2023mgfn} combined magnitude-contrastive learning with glance-and-focus mechanisms for video anomaly detection. \cite{sultani2018real} focused on real-time anomaly detection, highlighting the necessity for efficient, real-time solutions.

\begin{table}[ht]
\centering
\caption{Overview of Quantum Anomaly Detection Algorithms.}

\resizebox{0.45\textwidth}{!}{
\begin{tabular}{p{3cm}p{4cm}p{4cm}p{2cm}p{1cm}}
\toprule
\toprule
\textbf{Model} & \textbf{Algorithm} &\textbf{Data Set} & \textbf{Supervised}  & \textbf{VQC} \\
\midrule
\cite{liu2018quantum} &  quantum version of the kernel PCA algorithm and  quantum one-class SVM & quantum data (quantum states) & \Ellipse & \XSolidBrush\\
\addlinespace[7pt]
\cite{guo2022quantum} &  Amplitude Estimation & N/A (Complexity Comparison) &\XSolidBrush &\XSolidBrush\\
\addlinespace[7pt]
\cite{kottmann2021variational} & VQC (only including RY and (Controlled-z gate)) & quantum data set & \XSolidBrush & \CheckmarkBold  \\
\addlinespace[7pt]
\cite{guo2023quantum} & Quantum Local Outlier Factor algorithm & N/A (Complexity Comparison) &\XSolidBrush & \XSolidBrush \\
\addlinespace[7pt]
\cite{wang2023quantum} & QHDNN only contains RY gates & MNIST and FashionMNIST &\CheckmarkBold &\CheckmarkBold \\
\addlinespace[7pt]
\cite{moro2023anomaly} & quantum-restricted Boltzmann machines & NSL-KDD and CSE-CIC-IDS2018 & \CheckmarkBold & \EllipseSolid\\
\addlinespace[7pt]
\hline
\addlinespace[7pt]
Ours (Qsco) & VQC based & Eight real-world anomaly detection datasets & \CheckmarkBold &\CheckmarkBold \\
\bottomrule
\bottomrule
\multicolumn{4}{l}{\Ellipse indicates semi-supervised and \EllipseSolid indicates both of VQC and quantum mechanics.}
\end{tabular}
}

\end{table}

\textbf{Quantum Anomaly Detection} Quantum anomaly detection utilizes principles of quantum computing to significantly enhance the identification of data outliers, which is crucial for fraud detection, network security, and medical diagnostics applications. This emerging field has seen promising developments: \cite{liu2018quantum} introduced quantum algorithms for kernel principal component analysis (PCA) and one-class support vector machine (SVM), which can operate with resources logarithmic in the dimensionality of quantum states. \cite{guo2022quantum} proposed an efficient quantum algorithm for the computationally demanding ADDE algorithm, using amplitude estimation to handle large datasets effectively. Similarly, \cite{kottmann2021variational} developed a variational quantum anomaly detection algorithm capable of extracting phase diagrams from quantum simulations without prior physical knowledge, fully operable on the quantum device used for the simulations. Extending further, \cite{guo2023quantum} transformed the Local Outlier Factor (LOF) algorithm into a quantum version, achieving exponential speedup on data dimensions and polynomial speedup on data quantity. Additionally,\cite{wang2023quantum} engineered a quantum-classical hybrid DNN (QHDNN) that learns from normal images to isolate anomalies, exploring various quantum layer architectures and implementing a VQC-based solution. Lastly, \cite{moro2023anomaly} explored the potential of adiabatic quantum annealers (AQAs) for quantum speed-up in anomaly detection. 

\section{Proposed Approach}

\textbf{Problem Statement} The studied problem, open-set supervised AD, can be formally stated as follows. Given a set of training samples $\mathcal{X} =\left \{ \textbf{x}_i \right \} _{i = 1}^{N+M} $, in which $\mathcal{X}_n = \left \{ \textbf{x}_1,\textbf{x}_2 ,\dots ,\textbf{x}_N \right \} $ is the normal sample set and $\mathcal{X}_a = \left \{ \textbf{x}_{N+1},\textbf{x}_{N+2} ,\dots ,\textbf{x}_{N+M} \right \} $ $( M\ll N )$ is a very small set of annotated anomalies that provide some knowledge about true anomalies, and the $M$ anomalies belong to the seen anomaly classes $\zeta  \subset \varpi $, where $\varpi =\left \{ c_i \right \} _{i=1}^{\left |\varpi  \right | } $ denotes the set of all possible anomaly classes, and then the goal is detect both seen and unseen anomaly classes by learning an anomaly scoring function $g :\mathcal{X} \rightarrow \mathbb{R}  $ that assigns larger anomaly scores to both seen and unseen anomalies than normal samples.

\textbf{Quantum Scoring Module} In open-set anomaly detection, the model needs to identify new categories or anomalies not seen during the training phase. This type of anomaly detection is critical for many applications, such as security monitoring, medical diagnostics, and industrial quality control, where new and unknown anomalies often arise. 

For an image in the dataset $\mathcal{X} =\left \{ \textbf{x}_i \right \} _{i = 1}^{N+M} $, $\textbf{x}_i \in \mathbb{R} ^{H \times W \times C} $. Usually, this image can be represented as a three-dimensional tensor, which for color images usually contains three color channels (red, green, blue). Let the height of the image be $H$, the width is $W$, and the color channel is $C$. 

The feature extractor $\mathcal{F}(\cdot)$ is a function that converts the input image into a feature vector or matrix. It may contain multiple convolution layers, pooling layers, and activation layers and can be expressed as:
\begin{equation}
   \mathcal{X'} = \mathcal{F}(\mathcal{X}): \mathbb{R} ^{H \times W \times C} \longrightarrow \mathbb{R} ^{H' \times W' \times D},
\end{equation}

where $D$ represents feature depth and $(D\gg C)$.

\begin{figure}[htbp]

\begin{center}
\centerline{\includegraphics[width=0.5\textwidth]{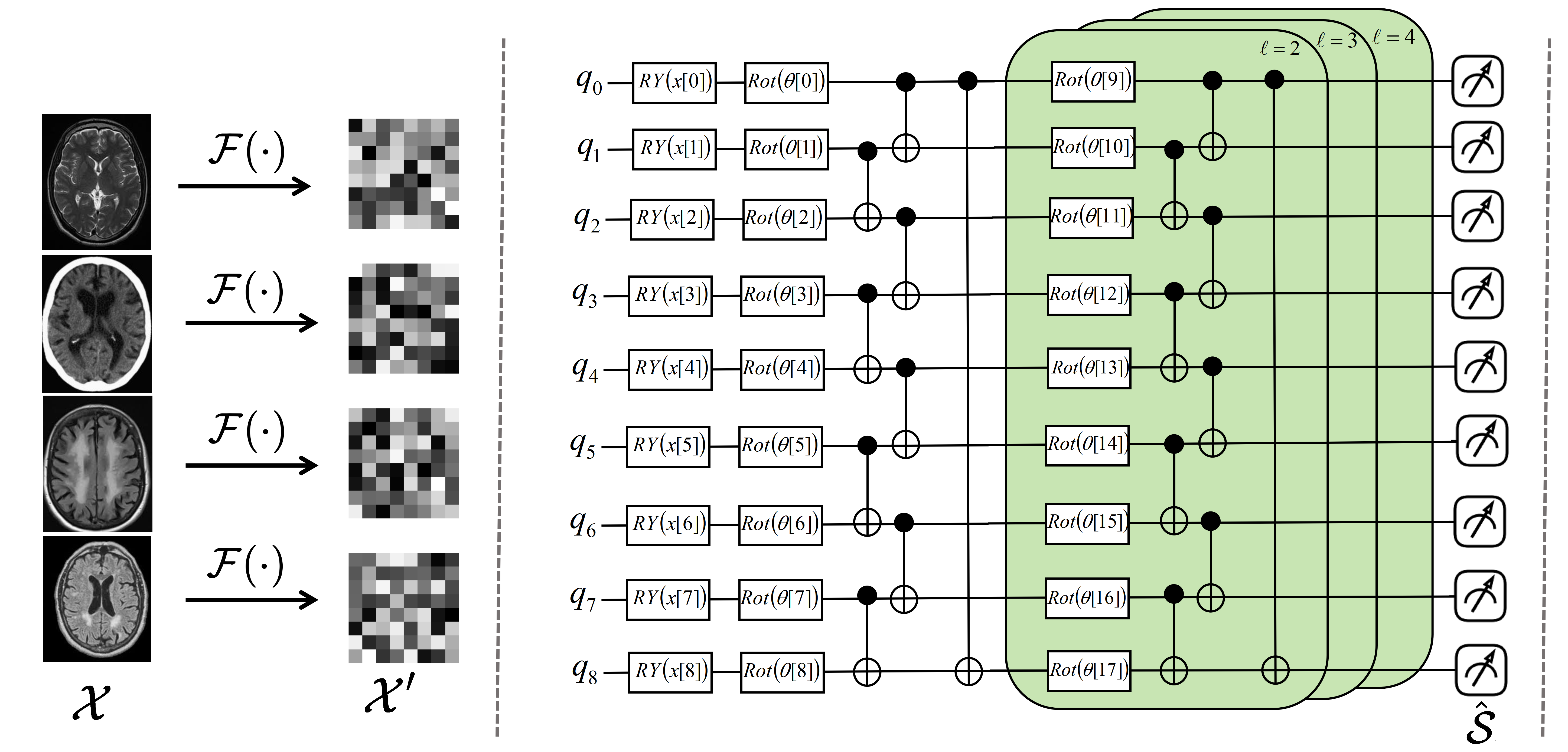}}
\caption{\textbf{Overview of our proposed Qsco module}. The left-hand section displays dataset samples from open set data, denoted as the input $\mathcal{X}$, which are fed into the feature extractor $\mathcal{F}\left ( \cdot  \right ) $ to obtain the feature map $\mathcal{X}'$. This feature map $\mathcal{X}'$ is the input to the quantum variational circuit. In the circuit, the $x$ values are used in the $RY(\cdot)$ gates, while the $\theta$ parameters are optimized during the training process in the $Rot(\cdot)$ operations. The green-shaded area represents the optimized layer in the Qsco module. The Qsco module output score is $\hat{\mathcal{S} }$.}

\label{quantumcircuit}

\end{center}
\end{figure}

The hyperbolic tangent function, or $\tanh$, is used to normalize input data to a fixed interval. For quantum computing's rotational gates (like RY), input typically needs to be in terms of angles, making tanh ideal for scaling data to the $\left [ -1,1 \right ] $ range, which helps control the rotation angles of these gates.
The next step is to apply the hyperbolic tangent activation function $\tanh$ to the feature vector $\mathcal{X'}$. The hyperbolic tangent function performs a nonlinear transformation on each element $\mathcal{X'}$, and each element of the output is $\left [ -1,1 \right ] $, mathematically expressed as :
\begin{equation}
    \mathcal{\hat{X}} = \tanh(\mathcal{X'})= \left [  \tanh(\textbf{x}'_{1}),\tanh(\textbf{x}'_{2}) ,\dots ,\tanh(\textbf{x}'_{N+ M})  \right ] .
\end{equation}



We applied a set of \(K\) qubits, denoted as \(\tau \in \mathbb{C}^K\), where each \(\tau_m\) represents the state of the \(m\)th qubit, in the Qsco module. A quantum state could be represented as:
\begin{equation}
    \left | \psi_m  \right \rangle =\alpha_m \left | 0  \right \rangle + \beta_m \left | 1  \right \rangle ,
\label{Qbas}
\end{equation}
where $\left | \psi_m  \right \rangle$ is the state of qubit, $\alpha_m$ and $\beta_m$ are the complex probability amplitudes, which satisfy: 
\begin{equation}
    \left | \alpha_m \right |^{2}  + \left | \beta_m \right |^{2} = 1.
  \label{a+b=1}  
\end{equation}
According to Eq. \ref{a+b=1}, Eq. \ref{Qbas} could be represented as:
\begin{equation}
    \left|\psi_m\right\rangle = e^{i\gamma_m} \left(\cos\frac{\theta_m}{2}\left|0\right\rangle + e^{i\varphi_m}\sin\frac{\theta_m}{2}\left|1\right\rangle\right),
\label{fullQbas}
\end{equation}
where \(\theta_m\), \(\varphi_m\), and \(\gamma_m\) are real numbers specific to the \(m\)th qubit. The factor \(e^{i\gamma_m}\) can be ignored due to its negligible effect \cite{nielsen2010quantum}. Therefore, we can rewrite Eq. \ref{fullQbas} as:
\begin{equation}
    \left|\psi_m\right\rangle = \cos\frac{\theta_m}{2}\left|0\right\rangle + e^{i\varphi_m}\sin\frac{\theta_m}{2}\left|1\right\rangle.
\end{equation}

The input of the quantum circuits is the $\hat{\mathcal{X}} = \left \{ \hat{\textbf{x}}_1,\hat{\textbf{x}}_2 ,\dots ,\hat{\textbf{x}}_{N+M} \right \}$ as shown in Fig. \ref{quantumcircuit}. We take $\hat{\textbf{x}}_m$ and utilize the Rotation Y gate (Ry gate) as below:
\begin{equation}
    Ry(\hat{\textbf{x}}_m) = \begin{pmatrix} \cos(\frac{\hat{\textbf{x}}_m}{2}) & -\sin(\frac{\hat{\textbf{x}}_m}{2}) \\ \sin(\frac{\hat{\textbf{x}}_m}{2}) & \cos(\frac{\hat{\textbf{x}}_m}{2}) \end{pmatrix} .
    \label{Rygate1}
\end{equation}

For a qubit in state in Eq. \ref{Qbas}, we can calculate the result as below:
\begin{equation}
    Ry(\hat{\textbf{x}}_m)|\psi_m\rangle =\begin{pmatrix} \alpha_m\cos(\frac{\hat{\textbf{x}}_m}{2})  -\beta_m\sin(\frac{\hat{\textbf{x}}_m}{2}) \\ \alpha_m\sin(\frac{\hat{\textbf{x}}_m}{2}) \beta_m \cos(\frac{\hat{\textbf{x}}_m}{2}) \end{pmatrix}.
\end{equation}

Then we initialize the parameters for the rotation gates $Rot\left ( \cdot  \right ) $, $\theta  \in \mathbb{R} ^{m \times n \times k} $ defined as below:
\begin{equation}
   Rot\left ( \cdot  \right ) = R_z(\lambda) R_y(\phi) R_x(\theta).
\end{equation}

\begin{figure}[ht]
\begin{center}
\centerline{\includegraphics[width=0.5\textwidth]{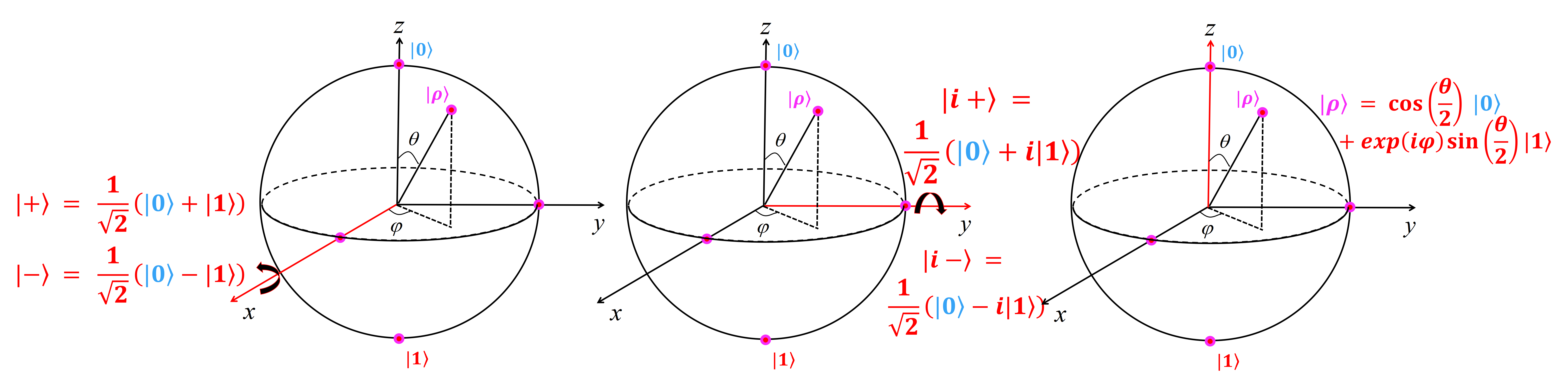}}
\caption{Visualization of the quantum state $\left | \rho  \right \rangle $ on the Bloch sphere during the quantum downsampling process. The left, middle, and right figures depict the effects of the single-qubit rotation gates $RX$, $RY$, and $RZ$.}
\label{FIG-AA}
\vspace{-15pt}
\end{center}
\end{figure}

Rx gate (Rotation around the X-axis gate) in quantum computing is used to rotate qubits around the X-axis of the Bloch ball, and the Ry gate is shown in Eq. \ref{Rygate1}. The Rotation Z gate (Rz gate) is used in quantum computing to rotate the qubit around the Z-axis as shown in Fig. \ref{FIG-AA}.
{\small
\begin{equation}
    Rx(\theta)= \begin{pmatrix} \cos(\frac{\theta}{2}) & -i\sin(\frac{\theta}{2}) \\ -i\sin(\frac{\theta}{2}) & \cos(\frac{\theta}{2}) \end{pmatrix} , Rz(\lambda)= \begin{pmatrix} e^{-i\frac{\lambda}{2}} & 0 \\ 0 & e^{i\frac{\lambda}{2}} \end{pmatrix}. 
\label{Rxgate}
\end{equation}
}
So the quantum state after the $Rot(\cdot)$ could be expressed in summary as:
\begin{equation}
     |\psi_m''\rangle = R_z(\lambda) R_y(\phi) R_x(\theta)  |\psi_m'\rangle
\end{equation}

The function of the CNOT gate is as follows: if the control bit is in the $\left | 1  \right \rangle $ state, it will perform a Pauli-X gate operation on the target bit (flip the state of the target bit). The target bit is unchanged if the control bit is in the $\left | 0  \right \rangle $ state. 

We utilize the CNOT gate to implement conditional logic between qubits:
\begin{equation}
   |\tilde{ \psi}_m\rangle =  |\psi_m''\rangle \otimes |\psi_{m+1}''\rangle, m \in \left [ 0,K -1  \right ] .
\end{equation}


We implement a Pauli-Z gate (Z gate for short) on a superposition state that will affect the phase of the qubit without changing its probability amplitude as:
\begin{equation}
    \sigma_z = \begin{bmatrix}
  1&0 \\
  0&-1
\end{bmatrix}.
\label{pauliZ}
\end{equation}
When we apply the Pauli-Z gate in Eq. \ref{pauliZ} to a qubit in Eq. \ref{Qbas}, we can get:
\begin{equation}
    \left | \psi  \right \rangle =\alpha \left | 0  \right \rangle - \beta \left | 1  \right \rangle .
\end{equation}
Physically, the Z-gate acts like a $180$ degree rotation of the qubit's phase. This phase rotation does not change the measured probability distribution of the qubit, but it affects the global phase of the quantum state:
\begin{equation}
   \tilde{\textbf{s}}_m =\sigma_z\cdot 
 |\tilde{ \psi}_{m}\rangle .
\end{equation}
The output from Qsco as the input from the feature extractor $\mathcal{\hat{X}}_n$ to the output $\mathcal{\hat{S}}_n$ is shown as below:
\begin{equation}
    \mathcal{\hat{X}} \longrightarrow 
 \mathcal{\hat{S}}, \mathcal{S} = \left \{ \textbf{s}_1,\textbf{s}_2 ,\dots ,\textbf{s}_{N+M} \right \}
\end{equation}
    
$ \mathcal{\hat{S}}$ is the output of the Qsco module, which can be added to the scoring part of any model.

\begin{figure}[htbp]

\begin{center}
\centerline{\includegraphics[width=0.45\textwidth]{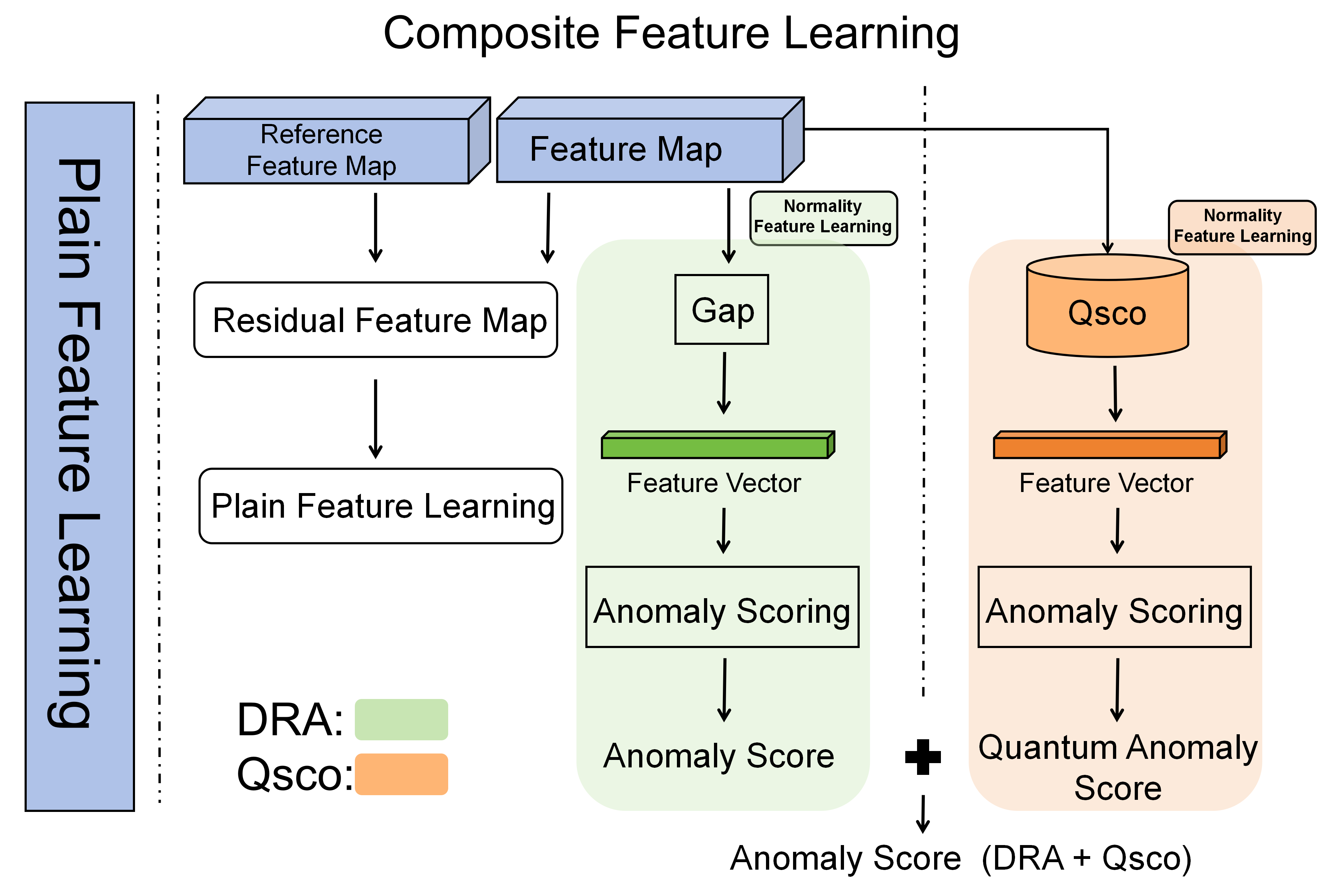}}

\caption{Overview of our proposed Qsco module and classical DRA \cite{ding2022catching} latent residual abnormality learning in a composite feature space.}

\label{draflow}
\vspace{-15pt}
\end{center}
\end{figure}

We chose DRA to be our backbone model because it is the SOTA method in the OSAD task, and we insert Qsco to DRA as an additional part to participate in the scoring process as shown in Fig. \ref{draflow}.

\begin{table}[ht]
\centering
\caption{AUC results (mean ± std) on 8 real-world AD datasets under \textbf{general} setting (noted by $\heartsuit$). The best and second-best results and the third-best are respectively highlighted in \textcolor{red}{\textbf{red}} and \textcolor{blue}{\textbf{blue}} and \textbf{bold}. $\varpi$ is the number of anomaly classes.}

\resizebox{0.45\textwidth}{!}{
\begin{tabular}{l|c|c|c|c}

\toprule
\toprule
\textbf{Dataset} $(\varpi) $   & \textbf{DRA} & \textbf{DRA + Qsco} ($\ell =2 $ )& \textbf{DRA + Qsco} ($\ell =1 $) & \textbf{DRA + Qsco} ($\ell =3 $)\\
\addlinespace[5pt]
\hline
\addlinespace[5pt]
\multicolumn{5}{c}{\textbf{Ten Training Anomaly Examples (Random)} $\heartsuit $ } \\
\hline
\addlinespace[5pt]
\textbf{AITEX} $(12)$ &  \textcolor{red} {$\bm{0.886_{\pm 0.021}}$}  &  $\bm{0.875_{\pm 0.029}} $ &  $0.832_{\pm 0.021} $ & \textcolor{blue} {$\bm{0.880_{\pm 0.020}} $ }\\

\addlinespace[5pt]
\textbf{SDD} $(1)$  &  $0.972_{\pm 0.010}$ & \textcolor{red}{ $ \bm{0.980_{\pm 0.007}}$ }& \textcolor{blue}{$\bm{0.979_{\pm 0.006}}$}  &  {$\bm{0.974_{\pm 0.010}}$} \\
\addlinespace[5pt]

\textbf{ELPV} $(2)$ &  \textcolor{red} { $\bm{0.821_{\pm 0.028}}$} & $0.819_{\pm 0.020}$ & \textcolor{blue} {$\bm{0.820_{\pm 0.017}}$} & $\bm{0.819_{\pm 0.018}}$\\

\addlinespace[5pt]

\textbf{Optical} $(1)$ & \textcolor{blue} {$\bm{0.967_{\pm 0.007}}$ }   &   \textcolor{red} {$\bm{0.969_{\pm 0.008}}$}  & $0.950_{\pm 0.014}$ &  $\bm{0.960_{\pm 0.009}}$ \\


\addlinespace[5pt]

\textbf{BrainMRI} $(1)$& { $0.959_{\pm 0.011}$ }  & \textcolor{blue}{ $\bm{0.972_{\pm 0.017}}$ } & \textcolor{red} {$\bm{0.978_{\pm 0.017}}$} & $\bm{0.961_{\pm 0.019}}$  \\

\addlinespace[5pt]

\textbf{HeadCT} $(1)$  & \textcolor{blue} {$\bm{0.998_{\pm 0.003}}$} & \textcolor{red} { $\bm{0.999_{\pm 0.003}}$ } & $\bm{0.989_{\pm 0.008}}$ & $0.980_{\pm 0.012}$ \\

\addlinespace[5pt]

\textbf{Hyper-Kvasir} $(4)$ &  \textcolor{red} {$\bm{0.840_{\pm 0.029}}$ }&  \textcolor{blue} {$\bm{0.813_{\pm 0.019}}$}  & $0.811_{\pm 0.025}$ &   $ \bm{0.812_{\pm 0.018}}$ \\

\addlinespace[5pt]

\textbf{MVTec AD}  $(-)$ &  {  $0.960_{\pm 0.011}$ }   &   \textcolor{red}{$\bm{0.968_{\pm 0.010}}$} & \textcolor{blue} {$\bm{0.966_{\pm 0.013}}$} & $\bm{0.961_{\pm 0.024}}$  \\

\addlinespace[5pt]

\multicolumn{5}{c}{\textbf{One Training Anomaly Example (Random)} $\heartsuit$ } \\ 

\hline

\addlinespace[5pt]

\textbf{AITEX} $(12)$ &  $ \bm{0.703_{\pm 0.038}}$  &  \textcolor{blue} {$\bm{0.708_{\pm 0.027}}$}   & ${0.685_{\pm 0.044}}$ &  \textcolor{red} { $\bm{0.728_{\pm 0.018}}$}\\

\addlinespace[5pt]
\textbf{SDD} $(1)$ &  {  $ \bm {0.853_{\pm 0.056}}$  }& \textcolor{red}{$\bm{ 0.876_{\pm 0.032}}$}  & $0.803_{\pm 0.064}$ &  \textcolor{blue} { $\bm{ 0.866_{\pm 0.067}}$}   \\ 

\addlinespace[5pt]

\textbf{ELPV} $(2)$ & \textcolor{blue}  {$ \bm{0.626_{\pm 0.074}}$} & \textcolor{red} { $\bm{0.749_{\pm 0.048}}$ }   & $0.588_{\pm 0.022}$ &  {$\bm{0.595_{\pm 0.038}}$} \\

\addlinespace[5pt]

\textbf{Optical} $(1)$ & \textcolor{blue} {  $\bm{0.894_{\pm 0.026}}$ }   & \textcolor{red} { $\bm{0.910_{\pm 0.011}}$ } & $0.859_{\pm 0.020}$ &  $\bm{0.879_{\pm 0.017}}$ \\


\addlinespace[5pt]

\textbf{BrainMRI} $(1)$ &  $0.638_{\pm 0.045}$   &  $ \bm{0.692_{\pm 0.051}}$ & \textcolor{red} {$ \bm{0.718_{\pm 0.043}}$} & \textcolor{blue} {$ \bm{0.708_{\pm 0.039}}$}  \\

\addlinespace[5pt]

\textbf{HeadCT} $(1)$ &   $\bm{0.804_{\pm 0.010}}$ &  
 $0.781_{\pm 0.007}$   & \textcolor{blue} {$\bm{0.818_{\pm 0.006}}$} & \textcolor{red} {$\bm{0.846_{\pm 0.003}}$} \\

\addlinespace[5pt]

\textbf{Hyper-Kvasir} $(4)$ &   $0.712_{\pm 0.010}$ &  {  $ \bm{0.768_{\pm 0.015}}$ } & \textcolor{red} {$ \bm{0.771_{\pm 0.008}}$}  & \textcolor{blue} {$ \bm{0.770_{\pm 0.014}}$ }  \\

\addlinespace[5pt]

\textbf{MVTec AD}  $(-)$ &   {  $0.904_{\pm 0.033}$ }   & \textcolor{red}  {$\bm{0.914_{\pm 0.029}}$}  &  $\bm{0.905_{\pm 0.031}}$ & \textcolor{blue} {$\bm{0.908_{\pm 0.035}}$}\\
\addlinespace[5pt]
 
\bottomrule

\bottomrule

\end{tabular}
}
\vspace{-10pt}
\label{AUCgeneralQlayer}
\end{table}

\section{Experimental Results}
\textbf{Hyper parameters} We employ Adam as the optimizer, initialized with a learning rate of $0.0002$ and a weight decay factor of $1e-5$. The StepLR scheduler is utilized with the learning rate decaying by $0.1$ every $10$ epoch. The batch size is $48$, and steps for each training epoch are $20$, and there are $30$ epochs for training. Many studies use synthetic anomaly detection datasets derived from well-known image classification benchmarks, like MNIST \cite{lecun2010mnist} and CIFAR-10 \cite{cifar10}, utilizing either one-vs-all or one-vs-one approaches. These methods create anomalies that are significantly different from standard samples. However, the differences between anomalies and standard samples are often subtle and minor in real-world scenarios, such as detecting defects in industrial settings or identifying lesions in medical imaging. Consequently, following previous research \cite{ding2022catching}, we prioritize natural anomaly datasets, including five industrial defect inspection datasets: MVTec AD \cite{Bergmann2019MVTecA}, AITEX  \cite{SilvestreBlanes2019APFAITEX}, SDD \cite{Tabernik_2019SDD}, ELPV \cite{Deitsch_2019ELPF} and Optical\cite{wieler_2023_8086136}, and three more real-world datasets: BrainMRI \cite{salehi2021multiresolutionheadct} and HeadCT \cite{salehi2021multiresolutionheadct}, Hyperkvasir \cite{borgli2020hyperkvasir}. Experiments are performed on an NVIDIA $3060$ GPU, $48$ GB RAM, and an NVIDIA RTX A6000, $48$ GB RAM. We built the Qsco quantum circuit on Pennylane \cite{pennylane} for quantum simulation.

\textbf{One and Ten Anomaly Examples} It means the number of anomaly data in the training set is one and ten, respectively. We use the following two experiment protocols:

\textbf{General setting} It mimics a typical open-set anomaly detection situation, where a few anomaly examples are randomly selected from all potential anomaly classes in the test set for each dataset. These selected anomalies are then excluded from the test data. This approach aims to reflect real-world conditions where it is uncertain which anomaly classes are known and the extent of the anomaly classes covered by the given examples. Consequently, the datasets may include both seen and unseen anomaly classes or solely the seen anomaly classes, depending on the complexity of the applications, such as the total number of possible anomaly classes.

\textbf{Hard setting} It aims to specifically assess the models' ability to detect unseen anomaly classes, which is a critical challenge in open-set AD. To achieve this, anomaly example sampling is restricted to a single anomaly class, and all samples from this class are removed from the test set to ensure it includes only unseen anomaly classes. It is important to note that this setting applies only to datasets with at least two anomaly classes.


In Table \ref{AUCgeneralQlayer}, we conducted an empirical study on how changing the depth of the Qsco model (the number of layers $\ell$) affects its performance across eight different datasets. We categorized the tests into scenarios involving ten training anomalies and one training anomaly. Our results imply that increasing the parameter $\ell$, which controls the depth of layers in Qsco, leads to better anomaly detection AUC scores, especially in scenarios with only one training anomaly. As $\ell$ increases, Qsco's variational circuit becomes deeper, enhancing its ability to detect anomalies in more challenging tasks (performance on one training anomaly is better on ten training anomalies). However, increasing the depth can reduce performance in simpler scenarios with ten anomalies in the training set, possibly due to over-fitting. This highlights the importance of selecting an optimal value for $\ell$, as the model's complexity needs to be adjusted based on the specific characteristics of the task. For MVTec AD, more class details (noted as $(-)$) can be found in \textbf{Supplementary Material} Table. 1.

\begin{table}[ht]
\centering

\caption{\textbf{Time complexity (s) per epoch}. $\varpi$ is the number of anomaly classes. The testing computing resource is an NVIDIA GeForce RTX 3060 and $64$GB RAM. Simulation is processed by Pennylane 0.35.1. $\heartsuit $ means general settings. {\color{red}\textbf{Red}} means the biggest time consumption.}

\resizebox{0.45\textwidth}{!}{
\begin{tabular}{|l|c|c|c|c|c|}

\toprule
\textbf{Dataset} $(\varpi) $&  \textbf{DevNet}  & \textbf{DRA}  & \textbf{DRA + Qsco} ($\ell =2 $)& \textbf{DRA + Qsco} ($\ell =1 $) & \textbf{DRA + Qsco} ($\ell =3 $)\\
\addlinespace[5pt]
\hline
\addlinespace[5pt]
\multicolumn{6}{c}{\textbf{Ten Training Anomaly Examples (Random) } $\heartsuit $ } \\
\hline
\addlinespace[5pt]
\textbf{AITEX} $(12)$ &  $20.93 _{\pm 0.76}$  &  $36.75_{\pm 0.50} $ &  $42.22_{\pm 2.01 } $ ($14.9\% \uparrow$) & $41.38_{\pm 1.02 } $ ($12.6\% \uparrow$) &   $43.53_{\pm 2.66 } $ ($\color{red}\bm{18.4}\% \uparrow$)\\

\addlinespace[5pt]
\textbf{SDD} $(1)$  &  $21.28 _{\pm 0.98}$  &  $36.59_{\pm 3.51} $ &  $42.28_{\pm 2.37 } $ ($15.6\% \uparrow$) & $40.00_{\pm 2.72 } $ ($9.3\% \uparrow$) &   $45.43_{\pm 3.15 } $ ($\color{red}\bm{24.2}\% \uparrow$) \\
\addlinespace[5pt]

\textbf{ELPV} $(2)$ &  $21.41 _{\pm 0.95}$  &  $34.30_{\pm 2.30} $ &  $41.79_{\pm 2.29 } $ ($21.8\% \uparrow$) & $40.80_{\pm 3.12 }$ ($19.0\% \uparrow$) & $44.31_{\pm 2.93 } $ ($\color{red}\bm{29.2}\% \uparrow$)\\

\addlinespace[5pt]

\textbf{Optical} $(1)$ &  $21.43_{\pm 1.04}$  &  $35.40_{\pm 1.55 } $ &  $42.81 _{\pm 3.06 } $ ($20.9\% \uparrow$) & $40.44_{\pm 2.26 } $ ($14.2\% \uparrow$)  &   $43.97 _{\pm 3.43 } $ ($\color{red}\bm{24.2}\% \uparrow$) \\


\addlinespace[5pt]

\textbf{BrainMRI} $(1)$&  $13.81_{\pm 0.72}$  &  $19.12 _{\pm 0.17} $ &  $24.25_{\pm 0.44 } $ ($26.8\% \uparrow$) & $ 22.85_{\pm 0.50 } $ ($19.5\% \uparrow$) & $26.45_{\pm 0.52 } $ ($\color{red}\bm{38.3}\% \uparrow$)\\

\addlinespace[5pt]

\textbf{HeadCT} $(1)$ &  $14.03_{\pm 0.55}$  &  $19.18_{\pm 0.25} $ &  $24.64_{\pm 0.37 } $ ($28.5\% \uparrow$) & $22.67_{\pm 0.27 } $ ($18.2\% \uparrow$) &   $26.26_{\pm 0.13 } $ ($\color{red}\bm{37.0}\% \uparrow$) \\

\addlinespace[5pt]

\textbf{Hyper-Kvasir} $(4)$ &  $21.47_{\pm 1.09}$  &  $34.06_{\pm 2.77} $ &  $41.70_{\pm 3.21 } $ ($22.4\% \uparrow$) & $41.37_{\pm 5.55 } $ ($21.5\% \uparrow$) & $53.65_{\pm 3.43 } $ ($\color{red}\bm{57.5}\% \uparrow$)\\

\addlinespace[5pt]

\multicolumn{6}{c}{\textbf{One Training Anomaly Example (Random)} $\heartsuit$ } \\ 

\hline

\addlinespace[5pt]

\textbf{AITEX} $(12)$ &  $20.89 _{\pm 1.07}$  &  $32.66_{\pm 2.09} $ &  $44.12_{\pm 1.86 } $ ($35.1\% \uparrow$)& $43.68_{\pm 4.56 } $ ($33.7\% \uparrow$) &   $44.74_{\pm 1.5 } $ ($\color{red}\bm{37.0}\% \uparrow$)\\

\addlinespace[5pt]
\textbf{SDD} $(1)$ &  $21.90 _{\pm 1.31}$  &  $35.14_{\pm 2.09} $ &  $40.54_{\pm 3.20 } $ ($15.4\% \uparrow$) & $38.64_{\pm 3.03 } $ ($10.0\% \uparrow$) & $42.06_{\pm 3.40 } $ ($\color{red}\bm{19.7}\% \uparrow$) \\

\addlinespace[5pt]

\textbf{ELPV} $(2)$ &  $21.39 _{\pm 1.80}$  &  $33.01_{\pm 2.15} $ &  $46.60_{\pm 4.03 } $ ($41.2\% \uparrow$) & $43.50_{\pm 2.85 } $ ($31.8\% \uparrow$) &   $47.33_{\pm 5.22 } $ ($\color{red}\bm{43.4}\% \uparrow$) \\

\addlinespace[5pt]

\textbf{Optical} $(1)$ &  $20.94 _{\pm 1.30}$  &  $34.01_{\pm 1.70} $ &  $42.92_{\pm 3.50 } $ ($26.2\% \uparrow$) & $40.84_{\pm 2.12 } $ ($20.1\% \uparrow$) &   $44.18_{\pm 0.35 } $ ($\color{red}\bm{30.0}\% \uparrow$)\\


\addlinespace[5pt]

\textbf{BrainMRI} $(1)$ &  $14.19 _{\pm 0.87}$  &  $19.33_{\pm 0.82} $ &  $24.46_{\pm 1.97 } $ ($26.5\% \uparrow$) & $23.95_{\pm 2.11 } $ ($24.0\% \uparrow$) &   $30.82_{\pm 4.07} $ ($\color{red}\bm{59.4}\% \uparrow$) \\

\addlinespace[5pt]

\textbf{HeadCT} $(1)$ &  $14.73_{\pm 0.71}$  &  $19.74_{\pm 0.79} $ &  $30.20_{\pm 1.67 } $ ($53.0\% \uparrow$) & $22.73_{\pm 1.65 } $ ($15.1\% \uparrow$) &   $31.09_{\pm 5.10 } $ ($\color{red}\bm{57.5}\% \uparrow$)\\

\addlinespace[5pt]

\textbf{Hyper-Kvasir} $(4)$  &  $23.08 _{\pm 1.22}$  &  $34.55_{\pm 2.08} $ &  $41.75_{\pm 3.29 } $ ($20.8\% \uparrow$) & $41.15_{\pm 4.28 } $ ($19.1\% \uparrow$) &   $47.70_{\pm 2.67 } $ ($\color{red} \bm{38.1}\% \uparrow$) \\

\addlinespace[5pt]

\bottomrule

\end{tabular}
}
\vspace{-15pt}
\label{Trainingtime}
\end{table}


Meanwhile, we measured the simulation time of quantum variational circuits using Pennylane 0.35.1, and the computing resource used was an NVIDIA GeForce RTX 3060 and 64GB RAM. While Qsco performance has been improved as indicated in Table \ref{AUCgeneralQlayer}, the quantum approach also has a relative shortcoming: it will cause a certain amount of time loss. However, from the table. \ref{Trainingtime}, we observe that the additional time consumption of approximately $20\%$ (6 seconds) is acceptable compared to the performance enhancement, attributed to the fewer quantum bits $\tau = 9$. The reason behind the time loss is the deepening of the quantum variational circuit, which ultimately leads to an increase in the revolving gate operations $Rot(\cdot)$. Nevertheless, a too-deep circuit (bigger $\ell$) will lead to over-fitting. On the contrary, a too-shallow circuit will be unable to explore the abnormal data space, leading to under-fitting problems. Therefore, we need to consider fewer quantum gate operations while ensuring performance because, in the NISQ era \cite{preskill2018quantum}, a controllable number of quantum gates has higher implement ability and operability.

\begin{table}[ht]
\centering
\caption{AUC results (mean ± std) on 8 real-world AD datasets under \textbf{general} setting (noted by $\heartsuit$). The best and second-best results and the third-best are respectively highlighted in \textcolor{red}{\textbf{red}} and \textcolor{blue}{\textbf{blue}} and \textbf{bold}. $\varpi$ is the number of anomaly classes. SAOE, FLOS, and MLEP results are from \citet{ding2022catching}. }

\resizebox{0.45\textwidth}{!}{
\begin{tabular}{l|c|c|c|c|c|c}

\toprule
\toprule
\textbf{Dataset} $(\varpi) $ & \textbf{SAOE}  & \textbf{FLOS} & \textbf{MLEP} & \textbf{DevNet} & \textbf{DRA} & \textbf{DRA + Qsco} ($\ell =2 $)  \\
\addlinespace[5pt]
\hline
\addlinespace[5pt]
\multicolumn{7}{c}{\textbf{Ten Training Anomaly Examples (Random)} $\heartsuit $ } \\
\hline
\addlinespace[5pt]
\textbf{AITEX} $(12)$ &  $0.874_{\pm 0.024}$ & $0.841_{\pm 0.049}$ & $0.867_{\pm 0.037}$ &  \textcolor{blue} {$\bm{0.881_{\pm 0.039}}$} &  \textcolor{red} {$\bm{0.886_{\pm 0.021}}$}  &  $\bm{0.875_{\pm 0.029}} $  \\

\addlinespace[5pt]
\textbf{SDD} $(1)$ &  $0.955_{\pm 0.020}$  &  $0.967_{\pm 0.018}$  &  $0.783_{\pm 0.013}$  &  $\bm{0.977_{\pm 0.010}}$  &  \textcolor{blue}{$\bm{0.972_{\pm 0.010}}$ } & \textcolor{red} { $ \bm{0.980_{\pm 0.001}}$ } \\
\addlinespace[5pt]

\textbf{ELPV} $(2)$ &  $0.793_{\pm 0.047}$ &  $0.818_{\pm 0.032}$  &  $0.794_{\pm 0.047}$ &   \textcolor{red} {$\bm{0.856_{\pm 0.013}}$ } & \textcolor{blue} { $\bm{0.821_{\pm 0.028}}$} &  $\bm{0.819_{\pm 0.020}}$ \\

\addlinespace[5pt]

\textbf{Optical} $(1)$ &  $\bm{0.941_{\pm 0.013}}$ &  $0.720_{\pm 0.055}$ &  $0.740_{\pm 0.039}$ &  $0.777_{\pm 0.022}$ & \textcolor{blue} {  $\bm{0.967_{\pm 0.007}}$ }   &   \textcolor{red} {$\bm{0.969_{\pm 0.008}}$}    \\


\addlinespace[5pt]

\textbf{BrainMRI} $(1)$&  $0.900_{\pm 0.041}$ &  $0.955_{\pm 0.011}$ &  $\bm{0.959_{\pm 0.011}}$  &  $0.959_{\pm 0.012}$ & \textcolor{blue} { $\bm{0.959_{\pm 0.011}}$ }  &  \textcolor{red} { $\bm{0.972_{\pm 0.017}}$ }   \\

\addlinespace[5pt]

\textbf{HeadCT} $(1)$ &  $0.935_{\pm 0.021}$ &  $0.971_{\pm 0.004}$  &  $0.972_{\pm 0.014}$ & $\bm{0.993_{\pm 0.004}}$     & \textcolor{blue} {  $\bm{0.998_{\pm 0.003}}$} &  
\textcolor{red} { $\bm{0.999_{\pm 0.003}}$ } \\

\addlinespace[5pt]

\textbf{Hyper-Kvasir} $(4)$ &  $0.666_{\pm 0.050}$  &  $0.773_{\pm 0.029}$ &  $0.600_{\pm 0.069}$ &  \textcolor{red} { $\bm{0.876_{\pm 0.012}}$ } &   \textcolor{blue} {$\bm{0.840_{\pm 0.029}}$ }&  $ \bm{0.813_{\pm 0.019}}$     \\

\addlinespace[5pt]

\textbf{MVTec AD}  $(-)$ &  $0.926_{\pm 0.010}$ &  $\bm{0.939_{\pm 0.007}}$  &  $0.907_{\pm 0.005}$ &  $0.934_{\pm 0.019}$ & \textcolor{blue} {  $\bm{0.960_{\pm 0.011}}$ }   &  \textcolor{red} {$\bm{0.968_{\pm 0.010}}$} \\

\addlinespace[5pt]

\multicolumn{7}{c}{\textbf{One Training Anomaly Example (Random)} $\heartsuit$ } \\ 

\hline

\addlinespace[5pt]

\textbf{AITEX} $(12)$ &  $0.675_{\pm 0.094}$ & $0.538_{\pm 0.073}$ & $0.564_{\pm 0.055}$ &  $\bm{0.627_{\pm 0.031}}$ &   \textcolor{blue} {$ \bm{0.703_{\pm 0.038}}$ } &  \textcolor{red} {$\bm{0.708_{\pm 0.027}}$}     \\

\addlinespace[5pt]
\textbf{SDD} $(1)$ &  $0.781_{\pm 0.009}$  &  $0.840_{\pm 0.043}$  &  $0.811_{\pm 0.045}$  &  \textcolor{blue} { $\bm{0.873_{\pm 0.055}}$ } &   $ \bm {0.853_{\pm 0.056}}$  & \textcolor{red} {$\bm{ 0.876_{\pm 0.032}}$}       \\ 

\addlinespace[5pt]

\textbf{ELPV} $(2)$ &  $\bm{0.635_{\pm 0.092}}$ &  $0.457_{\pm 0.056}$  &  $0.578_{\pm 0.062}$ &  \textcolor{blue} {$\bm{0.691_{\pm 0.069}}$ } &  $0.626_{\pm 0.074}$ &  \textcolor{red} { $\bm{0.749_{\pm 0.048}}$ }     \\

\addlinespace[5pt]

\textbf{Optical} $(1)$ &  $\bm{0.815_{\pm 0.014}}$ &  $0.518_{\pm 0.003}$ &  $0.516_{\pm 0.009}$ &  $0.537_{\pm 0.016}$ & \textcolor{blue} {  $\bm{0.894_{\pm 0.026}}$ }   & \textcolor{red} { $\bm{0.910_{\pm 0.011}}$ }\\


\addlinespace[5pt]

\textbf{BrainMRI} $(1)$&  $0.531_{\pm 0.060}$ &  $0.693_{\pm 0.036}$ &  $0.632_{\pm 0.017}$  &  \textcolor{red} {$\bm{0.765_{\pm 0.030}}$ }&  $\bm{0.638_{\pm 0.045}}$   & \textcolor{blue} { $ \bm{0.718_{\pm 0.043}}$ }  \\

\addlinespace[5pt]

\textbf{HeadCT} $(1)$ &  $0.597_{\pm 0.022}$ &  $0.698_{\pm 0.092}$  &  $0.758_{\pm 0.038}$ &  $\bm{0.768_{\pm 0.031}}$     & \textcolor{red} { $\bm{0.804_{\pm 0.010}}$ }&  
 \textcolor{blue} { $\bm{0.781_{\pm 0.007}}$ }   \\

\addlinespace[5pt]

\textbf{Hyper-Kvasir} $(4)$ &  $0.498_{\pm 0.100}$  &  $0.668_{\pm 0.004}$ &  $0.445_{\pm 0.040}$ &  $\bm{0.682_{\pm 0.038}}$ & \textcolor{blue} { $\bm{0.712_{\pm 0.010}}$ }& \textcolor{red} {  $ \bm{0.768_{\pm 0.015}}$ }    \\

\addlinespace[5pt]

\textbf{MVTec AD}  $(-)$ &  $\bm{0.834_{\pm 0.007}}$ &  $0.792_{\pm 0.014}$  &  $0.744_{\pm 0.019}$ &  $0.812_{\pm 0.049}$ & \textcolor{blue} {  $\bm{0.904_{\pm 0.033}}$ }   &  \textcolor{red} {$\bm{0.914_{\pm 0.029}}$}  \\
\addlinespace[5pt]
 
\bottomrule

\bottomrule

\end{tabular}
}
\label{AUCgeneral}
\end{table}

In Table. \ref{AUCgeneral}, after applying the best-performing parameter $\ell = 2 $, we compared the performance (AUC score) with more baseline methods, namely SAOE (combining data augmentation-based Synthetic Anomalies \cite{SAOE1, SAOE2, SAOE3}, FLOS \cite{Flos}, MLEP \cite{MLEP}, and DevNet \cite{pang2021explainable,pang2019deep}. This table indicates that Qsco has better anomaly detection performance than most baseline methods among all datasets.

\begin{table}[ht]
\centering
\caption{AUC results (mean ± std) on 8 real-world AD datasets under the \textbf{hard} setting (noted by $\diamondsuit$). The best and second-best results and the third-best are respectively highlighted in \textcolor{red}{\textbf{red}} and \textcolor{blue}{\textbf{blue}} and \textbf{bold}. $\infty $ is the average of anomaly classes. SAOE, FLOS, MLEP, and DevNet results are from \citet{ding2022catching}.}

\resizebox{0.45\textwidth}{!}{
\begin{tabular}{l|c|c|c|c|c|c}

\toprule
\toprule
\textbf{Dataset}  & \textbf{SAOE}  & \textbf{FLOS} & \textbf{MLEP} & \textbf{DevNet} & \textbf{DRA} & \textbf{DRA + Qsco} ($\ell =2 $) \\
\addlinespace[5pt]
\hline
\addlinespace[5pt]
\multicolumn{7}{c}{\textbf{Ten Training Anomaly Examples (Random)} $\diamondsuit  $} \\
\hline
\addlinespace[5pt]
\textbf{Carpet} $\infty $ &  $0.762_{\pm 0.073}$ & $0.761_{\pm 0.012}$ & $0.751_{\pm 0.023}$ &  $\bm{0.847_{\pm 0.017}}$ & \textcolor{blue} { $\bm{0.920_{\pm 0.039}}$ }  &  \textcolor{red} {$\bm{0.927_{\pm 0.044}}$}    \\

\addlinespace[5pt]
\textbf{Metal-nut} $\infty $ &  $0.855_{\pm 0.016}$ & $0.922_{\pm 0.014}$ & $0.878_{\pm 0.058}$ & \textcolor{red} { $\bm{0.965_{\pm 0.011}}$ }&  $\bm{0.949_{\pm 0.021}}$  &  \textcolor{blue} {$\bm{0.956_{\pm 0.020}}$}  \\

\addlinespace[5pt]

\textbf{AITEX} $\infty $ &  \textcolor{red} {$\bm{0.724_{\pm 0.032}}$ }&  $0.635_{\pm 0.043}$  &  $0.626_{\pm 0.041}$ &  $0.683_{\pm 0.032}$  &  $\bm{ 0.700_{\pm 0.049}}$ &   \textcolor{blue} { $\bm{0.709_{\pm 0.038}}$ } \\

\addlinespace[5pt]

\textbf{ELPV} $\infty $ &  $0.683_{\pm 0.047}$ &  $0.646_{\pm 0.032}$ & \textcolor{red} {  $\bm{0.745_{\pm 0.020}}$ }&  $0.702_{\pm 0.023}$    &   $ \bm{0.705_{\pm 0.050}}$ & \textcolor{blue} {$\bm{0.731_{\pm 0.042}}$}  \\


\addlinespace[5pt]

\textbf{Hyper-Kvasir} $\infty $ &  $0.698_{\pm 0.021}$  &   \textcolor{blue} {$\bm{0.786_{\pm 0.021}}$ } &$ 0.571_{\pm 0.014}$ &  \textcolor{red} { $\bm{0.822_{\pm 0.019}}$ }&   $\bm{0.764_{\pm 0.047}}$ &  $0.750_{\pm 0.031}$   \\

\addlinespace[5pt]

\multicolumn{7}{c}{\textbf{One Training Anomaly Example (Random)} $\diamondsuit $} \\ 
\hline

\addlinespace[5pt]
\textbf{Carpet} $\infty $ &  $0.753_{\pm 0.055}$ & $0.678_{\pm 0.040}$ & $0.679_{\pm 0.029}$ &  $\bm{0.767_{\pm 0.018}}$ &  \textcolor{blue} {$\bm{0.878_{\pm 0.066}}$ } &  \textcolor{red} {$\bm{0.881_{\pm 0.072}}$}  \\

\addlinespace[5pt]
\textbf{Metal-nut} $\infty $ &  $0.816_{\pm 0.029}$ & $0.855_{\pm 0.024}$ & $0.825_{\pm 0.023}$ &  $\bm{0.855_{\pm 0.016}}$ &  \textcolor{red} {$\bm{0.932_{\pm 0.028}}$}  &  \textcolor{blue} {$\bm{0.928_{\pm 0.037}}$ }  \\

\addlinespace[5pt]

\textbf{AITEX} $\infty $ &  $ \bm{0.674_{\pm 0.034}}$ &  $0.624_{\pm 0.024}$  &  $0.466_{\pm 0.030}$ &  $0.646_{\pm 0.034}$  & \textcolor{red} {  $\bm{0.678_{\pm 0.071}}$} &    \textcolor{blue} {$\bm{0.676_{\pm 0.040}}$ }   \\

\addlinespace[5pt]

\textbf{ELPV} $\infty $ &  $0.614_{\pm 0.048}$ &  \textcolor{red} {$\bm{0.691_{\pm 0.008}}$ }&  $0.566_{\pm 0.111}$ &   \textcolor{blue} { $\bm{0.648_{\pm 0.057}}$} & $\bm{0.627_{\pm 0.056}}$    &  $0.616_{\pm 0.032}$    \\


\addlinespace[5pt]

\textbf{Hyper-Kvasir} $\infty $ &  $0.406_{\pm 0.018}$  &  $0.571_{\pm 0.004}$ &  $0.480_{\pm 0.044}$ &  $\bm{0.595_{\pm 0.023}}$ &  \textcolor{blue} {  $\bm{0.677_{\pm 0.065}}$ }& \textcolor{red} { $ \bm{0.687_{\pm 0.037}}$  }  \\

\addlinespace[5pt]

\bottomrule

\bottomrule

\end{tabular}
}
\vspace{-10pt}
\label{AUChard}
\end{table}


Table. \ref{AUChard} presents performance quantification following the established general configuration under ten and one anomaly examples. This table computes the average values for each dataset, with additional details available (subset for each dataset) in the \textbf{\textbf{Supplementary Material}} Table. 2  for ten anomaly examples and \textbf{\textbf{Supplementary Material}} Table. 3  for one anomaly example. Table. \ref{AUChard} demonstrates that for the dataset under "One Training Anomaly Examples (Random)," the introduction of Qsco significantly improves the performance metrics of the DRA. However, for "Ten Training Anomaly Example (Random)," it is important to recognize that integrating DRA and Qsco does not consistently achieve superior outcomes across the datasets. The proposed Qsco improves the performance of DRA under hard settings for ten training anomaly examples. However, DRA + Qsco doesn't always take the \textcolor{red}{\textbf{best}} performance. This discrepancy arises because the foundational DRA often demonstrates diminished effectiveness compared to alternative methods.  Therefore, the combined performance of DRA and Qsco falls short. This shortfall is primarily attributable to the inherent limitations of DRA, despite our efforts to enhance its performance.

\begin{figure}[ht]

\begin{center}
\centerline{\includegraphics[width=0.5\textwidth]{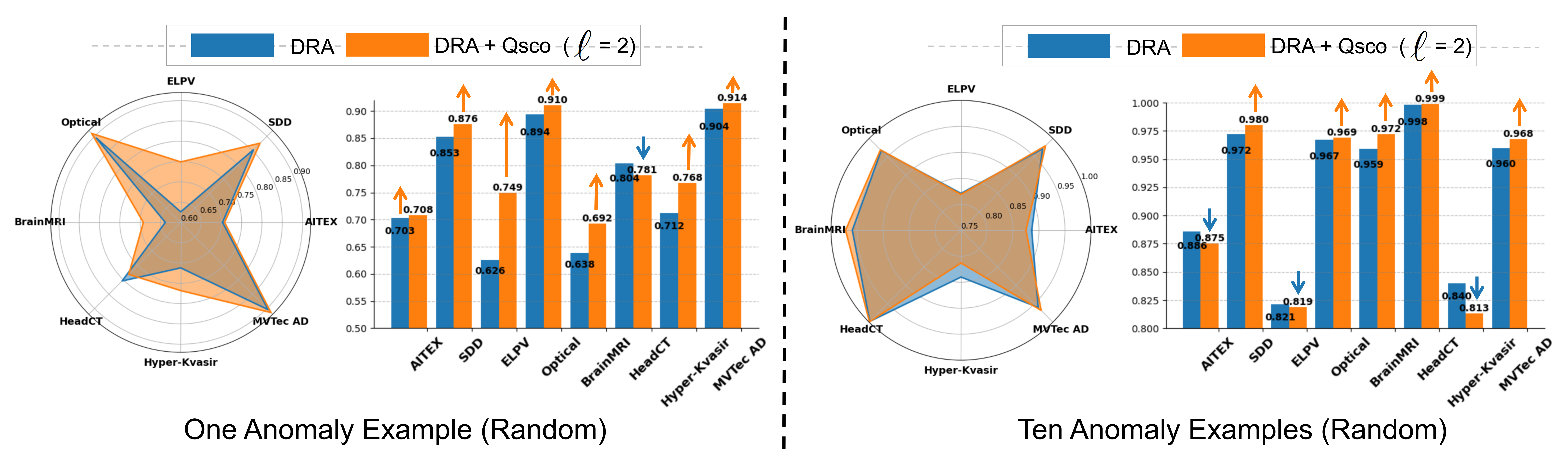}}
\caption{\textbf{Performance Comparison of DRA and Qsco ($\bm{\ell = 2}$) with different numbers of anomaly examples}. This comparison evaluates the performance (AUC score) of DRA and DRA + Qsco $\ell = 2$ under the same \textbf{general} settings across eight different datasets. The left side displays results for a single anomaly example, while the right shows results for ten. In the column chart, the ${\color{orange}\uparrow}$ indicates an improvement from the DRA model, and the ${\color{blue}\downarrow}$ signifies a decrease from the DRA.}
\label{radar110}
\vspace{-20pt}
\end{center}
\end{figure}


We present a comprehensive visualization of the performance metrics of Qsco ($\ell = 2$) across eight datasets, contrasting the results for scenarios involving one anomaly and ten anomaly examples in Fig. \ref{radar110}. Details for more $\ell$ could be found in \textbf{Supplementary Material} Fig.1 and Fig.2. The results reveal that Qsco significantly enhances the effectiveness of the DRA across a broad spectrum of evaluation metrics in situations under one anomaly example. This enhancement is consistent and marked by a notably more substantial improvement than scenarios with ten training anomaly examples. Furthermore, despite the inherently lower task complexity associated with ten training anomaly examples, Qsco demonstrates distinct advantages. The performance improvements, although less dramatic than in single anomaly scenarios, suggest that Qsco's methodology adapts effectively to varying degrees of complexity within anomaly detection tasks. This adaptability indicates robustness, critical for practical applications where anomaly conditions can vary widely.

\begin{table*}[h]
\centering
\caption{AUC results (mean ± std) on 7 real-world AD datasets under \textbf{general} setting (noted by $\heartsuit$). The increasing is highlighted from \colorbox{green!20}{low} to \colorbox{green!80}{high} and the decreasing (absolute values) is highlighted from \colorbox{red!20}{low} to \colorbox{red!80}{high} compared with $\ell =2 $. $\varpi$ is the number of anomaly classes. If the performance under the noise model is still better than the baseline (DRA), it is noted as $\flat$. The amplitude damping noise model is $\clubsuit$, and phase flipping noise is $\spadesuit $, depolarizing noise is $\ast $, and bit flipping noise is $\emptyset $.}

\resizebox{1\textwidth}{!}{
\begin{tabular}{l|c|c|c|c|c|c}

\toprule
\toprule
\textbf{Dataset} $(\varpi) $  & \textbf{DRA} & \textbf{DRA + Qsco} ($\ell =2 $ )& \textbf{DRA + Qsco} ($\ell =2 $) $\clubsuit$ & \textbf{DRA + Qsco} ($\ell =2 $) $\spadesuit $& \textbf{DRA + Qsco} ($\ell =2 $) $\ast $& \textbf{DRA + Qsco} ($\ell =2 $) $\emptyset $\\
\addlinespace[5pt]
\hline
\addlinespace[5pt]
\multicolumn{7}{c}{\textbf{Ten Training Anomaly Examples (Random)} $\heartsuit $ } \\
\hline
\addlinespace[5pt]
\textbf{AITEX} $(12)$ &  $\bm{0.886_{\pm 0.021}}$ &$\bm{0.875_{\pm 0.029}} $ &  \colorbox{red!30} {$\bm{0.869_{\pm 0.040}} $ $ (0.69\%)\downarrow$ }&  \colorbox{red!80} {$\bm{0.778_{\pm 0.032}} $ $ (11.09\%)\downarrow$ } &  \colorbox{red!62} {$\bm{0.801_{\pm 0.0334}} $ $ (8.46\%)\downarrow$ } &  \colorbox{red!70} {$\bm{0.794_{\pm 0.020}} $ $ (9.26\%)\downarrow$ } \\

\addlinespace[5pt]
\textbf{SDD} $(1)$& $\bm{0.972_{\pm 0.010}}$ & $ \bm{0.980_{\pm 0.007}}$ & \colorbox{red!40} {$\bm{0.966_{\pm 0.020}}$ $ (1.43\%)\downarrow$} & \colorbox{red!32} { $\bm{0.971_{\pm 0.006}}$ $ (0.92\%)\downarrow$} & \colorbox{red!34} {$\bm{0.968_{\pm 0.010}}$ $ (1.22\%)\downarrow$} &  \colorbox{red!27}{
$\bm{0.972_{\pm 0.012}}$ $ (0.82\%)\downarrow$} $\flat$\\
\addlinespace[5pt]

\textbf{ELPV} $(2)$ & $\bm{0.821_{\pm 0.028}}$ &$\bm{0.819_{\pm 0.020}}$ &  \colorbox{green!30} {$\bm{0.832_{\pm 0.021}}$ $ (1.59\%)\uparrow$} $\flat$& \colorbox{green!25} {$\bm{0.823_{\pm 0.018}}$$ (0.49\%)\uparrow$}$\flat$ &  \colorbox{green!30} {$\bm{0.832_{\pm 0.015}}$ $ (1.59\%)\uparrow$} $\flat$& \colorbox{green!38} { $\bm{0.840_{\pm 0.020}}$ $ (2.56\%)\uparrow$}$\flat$\\

\addlinespace[5pt]

\textbf{Optical} $(1)$ & $\bm{0.967_{\pm 0.007}}$  &$\bm{0.969_{\pm 0.008}}$  & \colorbox{green!20} {$\bm{0.971_{\pm 0.012}}$ $ (0.21\%)\uparrow$} $\flat$& \colorbox{green!15} { $\bm{0.970_{\pm 0.009}}$ $ (0.10\%)\uparrow$ } $\flat$& \colorbox{green!25} {$\bm{0.973_{\pm 0.017}}$ $ (0.41\%)\uparrow$} $\flat$& \colorbox{green!32} {$\bm{0.983_{\pm 0.005}}$ $ (1.44\%)\uparrow$}$\flat$\\


\addlinespace[5pt]

\textbf{BrainMRI} $(1)$& $\bm{0.959_{\pm 0.011}}$  &$\bm{0.972_{\pm 0.017}}$  & \colorbox{red!70} {$\bm{0.878_{\pm 0.018}} $ $(9.67\%) \downarrow$}  & \colorbox{red!60} {$\bm{0.893_{\pm 0.019}}$ $(8.13\%) \downarrow$ } & \colorbox{red!78} {$\bm{0.866_{\pm 0.020}} $ $(10.90\%) \downarrow$} & \colorbox{red!75} { $\bm{0.871_{\pm 0.022}}$   $(10.39\%) \downarrow$}\\

\addlinespace[5pt]

\textbf{HeadCT} $(1)$&$\bm{0.998_{\pm 0.003}}$  &  $\bm{0.999_{\pm 0.003}}$  & \colorbox{red!20} {$\bm{0.998_{\pm 0.002}}$ $ (0.10\%) \downarrow$} $\flat$ & \colorbox{red!25} {$0.990_{\pm 0.007}$ $ (0.90\%) \downarrow$} & \colorbox{red!45} {$\bm{0.982_{\pm 0.010}}$ $ (1.70\%) \downarrow$} &  \colorbox{red!45} {$\bm{0.982_{\pm 0.010}}$ $ (1.70\%) \downarrow$}\\

\addlinespace[5pt]

\textbf{Hyper-Kvasir} $(4)$ &$\bm{0.840_{\pm 0.029}}$ &$\bm{0.813_{\pm 0.019}}$  & \colorbox{green!50} {$\bm{0.852_{\pm 0.023}}$ $ (4.80\%) \uparrow$} $\flat$ &  \colorbox{red!45} {$0.789_{\pm 0.041}$ $ (2.95\%) \downarrow$} & \colorbox{red!70} {$\bm{0.740_{\pm 0.039}}$ $ (9.00\%) \downarrow$} & \colorbox{red!50} {$\bm{0.787_{\pm 0.038}}$  $ (3.20\%) \downarrow$}\\

\addlinespace[5pt]

\multicolumn{7}{c}{\textbf{One Training Anomaly Example (Random)} $\heartsuit$ } \\ 

\hline

\addlinespace[5pt]

\textbf{AITEX} $(12)$ & $ \bm{0.703_{\pm 0.038}}$  &$\bm{0.708_{\pm 0.027}}$   & \colorbox{green!40} {$\bm{0.734_{\pm 0.023}}$ $ (3.67\%) \uparrow$} $\flat$&  \colorbox{red!36} { $\bm{0.700_{\pm 0.038}}$ $ (1.13\%) \downarrow$ } &  \colorbox{green!25} {$\bm{0.722_{\pm 0.019}}$ $ (2.00\%) \uparrow$}$\flat$ &  \colorbox{green!10}{$\bm{0.708_{\pm 0.034}}$ $\longleftrightarrow$}$\flat$\\

\addlinespace[5pt]
\textbf{SDD} $(1)$ & $ \bm {0.853_{\pm 0.056}}$  &$\bm{ 0.876_{\pm 0.032}}$  & \colorbox{red!50} {$\bm{0.848_{\pm 0.045}}$ $ (3.20\%) \downarrow$} &   \colorbox{green!40} {$\bm{ 0.908_{\pm 0.026}}$  $ (3.65\%) \uparrow$} $\flat$& \colorbox{red!57} {$\bm{0.839_{\pm 0.045}}$ $ (4.22\%) \downarrow$} & \colorbox{green!34} {$\bm{ 0.904_{\pm 0.036}}$ $ (3.20\%) \uparrow$} $\flat$\\ 

\addlinespace[5pt]

\textbf{ELPV} $(2)$ & $\bm{0.626_{\pm 0.074}}$ &$\bm{0.749_{\pm 0.048}}$    & \colorbox{red!80} {$\bm{0.621_{\pm 0.046}}$ $ (17.10\%) \downarrow$ } $\flat$ &  \colorbox{green!10}{$\bm{0.749_{\pm 0.020}}$ $\longleftrightarrow$}$\flat$ &   \colorbox{green!34} {$\bm{ 0.770_{\pm 0.035}}$  $ (2.80\%) \uparrow$} $\flat$&   \colorbox{green!19} {$\bm{0.760_{\pm 0.031}}$ $ (1.47\%) \uparrow$}$\flat$ \\

\addlinespace[5pt]

\textbf{Optical} $(1)$ & $\bm{0.894_{\pm 0.026}}$ & $\bm{0.910_{\pm 0.011}}$  & \colorbox{red!45} {$\bm{0.893_{\pm 0.024}}$ $ (1.87\%) \downarrow$} & \colorbox{red!30} { $\bm{0.904_{\pm 0.015}}$ $ (0.66\%) \downarrow$ } $\flat$& \colorbox{red!50} {$\bm{0.890_{\pm 0.017}}$ $ (2.20\%) \downarrow$} &  \colorbox{red!17} { $\bm{0.909_{\pm 0.017}}$ $ (0.11\%)\downarrow$}$\flat$\\


\addlinespace[5pt]

\textbf{BrainMRI} $(1)$ & $\bm{0.638_{\pm 0.045}}$&$ \bm{0.718_{\pm 0.043}}$ & \colorbox{green!35} {$ \bm{0.739_{\pm 0.046}}$ $ (2.92\%) \uparrow$} $\flat$& \colorbox{green!40} {$ \bm{0.742_{\pm 0.040}}$ $ (3.34\%) \uparrow$} $\flat$& \colorbox{green!20} {$ \bm{0.729_{\pm 0.044}}$ $ (1.53\%) \uparrow$} $\flat$& \colorbox{red!45} {$ \bm{0.703_{\pm 0.048}}$ $ (2.09\%) \downarrow$}$\flat$\\

\addlinespace[5pt]

\textbf{HeadCT} $(1)$ & $\bm{0.804_{\pm 0.010}}$ &$\bm{0.781_{\pm 0.007}}$   &  \colorbox{green!80} {$\bm{0.871_{\pm 0.061}}$ $ (11.52\%) \uparrow$} $\flat$&\colorbox{green!70} { $\bm{0.847_{\pm 0.048}}$ $ (8.45\%) \uparrow$}$\flat$ & \colorbox{red!75} { $\bm{0.710_{\pm 0.100}}$ $ (9.10\%) \downarrow$ } &\colorbox{red!70} { $\bm{0.716_{\pm 0.085}}$ $ (8.32\%) \downarrow$}\\

\addlinespace[5pt]

\textbf{Hyper-Kvasir} $(4)$ &    $\bm{0.712_{\pm 0.010}}$   &$ \bm{0.768_{\pm 0.015}}$  & \colorbox{green!50} {$ \bm{0.804_{\pm 0.057}}$ $ (4.69\%) \uparrow$} $\flat$&  \colorbox{green!12} {$ \bm{0.771_{\pm 0.037}}$ $ (0.39\%) \uparrow$}  $\flat$& \colorbox{green!50} {$ \bm{0.804_{\pm 0.035}}$ $ (4.69\%) \uparrow$} $\flat$&   \colorbox{green!30} {$ \bm{0.788_{\pm 0.066}}$ $ (2.60\%) \uparrow$} $\flat$\\

\addlinespace[5pt]
 
\bottomrule

\bottomrule

\end{tabular}
}

\label{AUCgeneralnoise}
\end{table*}

\subsection{Noise Analysis}
In the NISQ era, anticipating the deployment of Qsco on actual quantum computing platforms necessitates addressing several critical issues, among which quantum noise is paramount. Accordingly, this section is dedicated to the application of various noise models to Qsco, enhancing our understanding of quantum noise in practical scenarios.

\textbf{Depolarizing noise.} The Depolarizing channel is denoted as $\Delta _\gamma$ with probability $\gamma$, replaces the state $\rho$ (density matrix) with the maximally mixed state $\frac{I}{d}$, where $I$ is the identity matrix and $d$ is the dimension of the Hilbert space.

The action of the depolarizing channel on a state $\rho$ is given by:
\begin{equation}
    \Delta_\gamma (\rho)=(1-\gamma )\rho+\frac{\gamma }{d}I.
\end{equation}

\textbf{Bit Flip Noise.} The bit flip noise flips the state from $\left | 0 \right \rangle $ to $\left |  1 \right \rangle $ or vice versa with probability $\gamma$ as
\begin{equation}
    \mathcal{E}(\rho) = (1-\gamma) \rho + \gamma X\rho X,
\end{equation}
where $X$ is the Pauli-X gate. 

\textbf{Phase Flip Noise.} Similar to the bit flipping noise. The probability $\gamma$, this noise applies a Pauli-Z gate to the quantum state, adding a negative sign to the $\left |  1 \right \rangle $ as:
\begin{equation}
    \mathcal{E}(\rho) = (1-\gamma) \rho + \gamma Z\rho Z.
\end{equation}

\textbf{Amplitude Damping.} The amplitude damping channel simulates the effect of energy dissipation from the qubit to the environment, decaying from the excited state  $\left |  1 \right \rangle $ to the ground state  $\left |  0 \right \rangle $. The effect can be represented using Kraus operators:
\begin{equation}
    \mathcal{E}(\rho) = E_0 \rho E_0^\dagger + E_1 \rho E_1^\dagger,
\end{equation}
where $E_0=\begin{pmatrix}
  1& 0\\
  0&\sqrt{1-\gamma } 
\end{pmatrix}$ and $E_1=\begin{pmatrix}
  0& \sqrt{\gamma }\\
  0& 0
\end{pmatrix}$ are the Kraus operators \cite{Nielsen_Chuang_2010}.

\textbf{Phase Damping.} Unlike amplitude damping, phase damping preserves the amplitude of the quantum state but loses phase information, describing a dephasing process of quantum information without energy loss as:
\begin{equation}
    \mathcal{E}(\rho) = E_0 \rho E_0^\dagger + E_1 \rho E_1^\dagger.
\end{equation}

As illustrated in Table. \ref{AUCgeneralnoise}, it is evident that various noise models adversely affect the AITEX, SDD, BrainMRI, HeadCT, and Hyper-Kvasir datasets in the context of Ten Training Anomaly Examples, with the overall maximum fluctuation around $10\%$. Conversely, performance on the ELPV and Optical datasets has improved. We hypothesize that introducing noise acts as a form of regularization for these two datasets, potentially preventing overfitting. This may allow the models to improve generalization and robustness in noisy real-world environments. For One and Ten Training Anomaly Examples in the Mvtecad dataset could be found in \textbf{\textbf{Supplementary Material}}. 

In the One Training Anomaly Example scenario, we observe an overall improvement, which is likely due to the reduced number of abnormal samples in the training data, which enables the model to focus more on capturing and learning the characteristics of normal data rather than being influenced by the noise associated with abnormal samples. Consequently, when noise is introduced, the model demonstrates a robust capacity to withstand minor data disturbances. This fortifies its adaptability and resilience against unseen noise, enhancing performance outcomes.

\begin{table}[ht]
\centering
\caption{Time complexity (s) on scalability analysis of Qsco.}
\resizebox{0.45\textwidth}{!}{
\begin{tabular}{c|c|c|c|c}
\hline
Qubits & 9 & 12 & 15 & 18\\ \hline
Pennylane        &       $13.11 \pm 0.25$       &      $17.26 \pm 1.37$     &   $58.55 \pm 1.24$ &   $182.19 \pm 2.35$ \\
\hline
\multicolumn{4}{l}{ Average of $100$ tests with NVIDIA 3060 with 48 GB RAM. }
\end{tabular}
}
\label{table4}

\end{table}

\textbf{Limitations and Future Work} The data presented in Table. \ref{table4} shows that classically simulating multiple instances of Qsco results in an exponential increase in time complexity. This exponential growth significantly constrains the scalability of classical simulations of Qsco. However, with the advent of frameworks such as NVIDIA's CUDA Quantum \cite{10247886}, which enables the simulation of quantum algorithms on CUDA-enabled GPUs, there is potential to enhance the scalability and efficiency of our experiments by leveraging GPU-accelerated quantum simulations in the future. Despite these limitations, current findings demonstrate that a single Qsco surpasses the established baseline in the OSAD task. Meanwhile, we will continue to compare with the most SOTA OSAD methods \cite{zhu2024anomaly}, striving to continuously improve Qsco and explore the potential of quantum computing in the OSAD domain.

\section{Conclusion}
This paper proposed a quantum scoring module Qsco for the OSAD task. This module can be embedded in any advanced model and learn the high-dimensional distribution of abnormal data through variational quantum circuits and entanglement between qubits. The proposed approach solves the problem of abnormal data not seen during the training process, which causes model performance degradation. Experimental results indicate that implementing Qsco markedly enhances the predicted AUC score and does not result in significant time loss. In this paper, we explored the application of quantum computing in anomaly detection by implementing Qsco. Drawing on results from a simulated quantum computing environment with noise models, we demonstrate a proof of concept highlighting the potential of integrating quantum computing and deep learning into open-set anomaly detection. 
\clearpage
\bibliography{aaai25}

\begin{thebibliography}{54}
\providecommand{\natexlab}[1]{#1}

\bibitem[{Abbaszade et~al.(2021)Abbaszade, Salari, Mousavi, Zomorodi, and Zhou}]{abbaszade2021application}
Abbaszade, M.; Salari, V.; Mousavi, S.~S.; Zomorodi, M.; and Zhou, X. 2021.
\newblock Application of quantum natural language processing for language translation.
\newblock \emph{IEEE Access}, 9: 130434--130448.

\bibitem[{Acsintoae et~al.(2022)Acsintoae, Florescu, Georgescu, Mare, Sumedrea, Ionescu, Khan, and Shah}]{acsintoae2022ubnormal}
Acsintoae, A.; Florescu, A.; Georgescu, M.-I.; Mare, T.; Sumedrea, P.; Ionescu, R.~T.; Khan, F.~S.; and Shah, M. 2022.
\newblock Ubnormal: New benchmark for supervised open-set video anomaly detection.
\newblock In \emph{Proceedings of the IEEE/CVF conference on computer vision and pattern recognition}, 20143--20153.

\bibitem[{Akcay, Atapour-Abarghouei, and Breckon(2019)}]{akcay2019ganomaly}
Akcay, S.; Atapour-Abarghouei, A.; and Breckon, T.~P. 2019.
\newblock Ganomaly: Semi-supervised anomaly detection via adversarial training.
\newblock In \emph{Computer Vision--ACCV 2018: 14th Asian Conference on Computer Vision, Perth, Australia, December 2--6, 2018, Revised Selected Papers, Part III 14}, 622--637. Springer.

\bibitem[{Alchieri et~al.(2021)Alchieri, Badalotti, Bonardi, and Bianco}]{alchieri2021introduction}
Alchieri, L.; Badalotti, D.; Bonardi, P.; and Bianco, S. 2021.
\newblock An introduction to quantum machine learning: from quantum logic to quantum deep learning.
\newblock \emph{Quantum Machine Intelligence}, 3(2): 28.

\bibitem[{Bergholm et~al.(2022)Bergholm, Izaac, Schuld, Gogolin, Ahmed, Ajith, Alam, Alonso-Linaje, AkashNarayanan, Asadi, Arrazola, Azad, Banning, Blank, Bromley, Cordier, Ceroni, Delgado, Matteo, Dusko, Garg, Guala, Hayes, Hill, Ijaz, Isacsson, Ittah, Jahangiri, Jain, Jiang, Khandelwal, Kottmann, Lang, Lee, Loke, Lowe, McKiernan, Meyer, Montañez-Barrera, Moyard, Niu, O'Riordan, Oud, Panigrahi, Park, Polatajko, Quesada, Roberts, Sá, Schoch, Shi, Shu, Sim, Singh, Strandberg, Soni, Száva, Thabet, Vargas-Hernández, Vincent, Vitucci, Weber, Wierichs, Wiersema, Willmann, Wong, Zhang, and Killoran}]{pennylane}
Bergholm, V.; Izaac, J.; Schuld, M.; Gogolin, C.; Ahmed, S.; Ajith, V.; Alam, M.~S.; Alonso-Linaje, G.; AkashNarayanan, B.; Asadi, A.; Arrazola, J.~M.; Azad, U.; Banning, S.; Blank, C.; Bromley, T.~R.; Cordier, B.~A.; Ceroni, J.; Delgado, A.; Matteo, O.~D.; Dusko, A.; Garg, T.; Guala, D.; Hayes, A.; Hill, R.; Ijaz, A.; Isacsson, T.; Ittah, D.; Jahangiri, S.; Jain, P.; Jiang, E.; Khandelwal, A.; Kottmann, K.; Lang, R.~A.; Lee, C.; Loke, T.; Lowe, A.; McKiernan, K.; Meyer, J.~J.; Montañez-Barrera, J.~A.; Moyard, R.; Niu, Z.; O'Riordan, L.~J.; Oud, S.; Panigrahi, A.; Park, C.-Y.; Polatajko, D.; Quesada, N.; Roberts, C.; Sá, N.; Schoch, I.; Shi, B.; Shu, S.; Sim, S.; Singh, A.; Strandberg, I.; Soni, J.; Száva, A.; Thabet, S.; Vargas-Hernández, R.~A.; Vincent, T.; Vitucci, N.; Weber, M.; Wierichs, D.; Wiersema, R.; Willmann, M.; Wong, V.; Zhang, S.; and Killoran, N. 2022.
\newblock PennyLane: Automatic differentiation of hybrid quantum-classical computations.
\newblock arXiv:1811.04968.

\bibitem[{Bergmann et~al.(2019)Bergmann, Fauser, Sattlegger, and Steger}]{Bergmann2019MVTecA}
Bergmann, P.; Fauser, M.; Sattlegger, D.; and Steger, C. 2019.
\newblock MVTec AD — A Comprehensive Real-World Dataset for Unsupervised Anomaly Detection.
\newblock \emph{2019 IEEE/CVF Conference on Computer Vision and Pattern Recognition (CVPR)}, 9584--9592.

\bibitem[{Borgli et~al.(2020)Borgli, Thambawita, Smedsrud, Hicks, Jha, Eskeland, Randel, Pogorelov, Lux, Nguyen et~al.}]{borgli2020hyperkvasir}
Borgli, H.; Thambawita, V.; Smedsrud, P.~H.; Hicks, S.; Jha, D.; Eskeland, S.~L.; Randel, K.~R.; Pogorelov, K.; Lux, M.; Nguyen, D. T.~D.; et~al. 2020.
\newblock HyperKvasir, a comprehensive multi-class image and video dataset for gastrointestinal endoscopy.
\newblock \emph{Scientific data}, 7(1): 283.

\bibitem[{Bozorgtabar and Mahapatra(2023)}]{bozorgtabar2023attention}
Bozorgtabar, B.; and Mahapatra, D. 2023.
\newblock Attention-conditioned augmentations for self-supervised anomaly detection and localization.
\newblock In \emph{Proceedings of the AAAI Conference on Artificial Intelligence}, volume~37, 14720--14728.

\bibitem[{Caro et~al.(2022)Caro, Huang, Cerezo, Sharma, Sornborger, Cincio, and Coles}]{caro2022generalization}
Caro, M.~C.; Huang, H.-Y.; Cerezo, M.; Sharma, K.; Sornborger, A.; Cincio, L.; and Coles, P.~J. 2022.
\newblock Generalization in quantum machine learning from few training data.
\newblock \emph{Nature communications}, 13(1): 4919.

\bibitem[{Chen et~al.(2023)Chen, Liu, Zhang, Fok, Qi, and Wu}]{chen2023mgfn}
Chen, Y.; Liu, Z.; Zhang, B.; Fok, W.; Qi, X.; and Wu, Y.-C. 2023.
\newblock Mgfn: Magnitude-contrastive glance-and-focus network for weakly-supervised video anomaly detection.
\newblock In \emph{Proceedings of the AAAI Conference on Artificial Intelligence}, volume~37, 387--395.

\bibitem[{Cho et~al.(2023)Cho, Kim, Hwang, Park, Lee, and Lee}]{cho2023look}
Cho, M.; Kim, M.; Hwang, S.; Park, C.; Lee, K.; and Lee, S. 2023.
\newblock Look around for anomalies: weakly-supervised anomaly detection via context-motion relational learning.
\newblock In \emph{Proceedings of the IEEE/CVF conference on computer vision and pattern recognition}, 12137--12146.

\bibitem[{Deitsch et~al.(2019)Deitsch, Christlein, Berger, Buerhop-Lutz, Maier, Gallwitz, and Riess}]{Deitsch_2019ELPF}
Deitsch, S.; Christlein, V.; Berger, S.; Buerhop-Lutz, C.; Maier, A.; Gallwitz, F.; and Riess, C. 2019.
\newblock Automatic classification of defective photovoltaic module cells in electroluminescence images.
\newblock \emph{Solar Energy}, 185: 455–468.

\bibitem[{Ding, Pang, and Shen(2022)}]{ding2022catching}
Ding, C.; Pang, G.; and Shen, C. 2022.
\newblock Catching Both Gray and Black Swans: Open-set Supervised Anomaly Detection.
\newblock In \emph{Proceedings of the IEEE/CVF Conference on Computer Vision and Pattern Recognition}.

\bibitem[{Farhi and Neven(2018)}]{Farhi2018ClassificationWQqnnqml}
Farhi, E.; and Neven, H. 2018.
\newblock Classification with Quantum Neural Networks on Near Term Processors.
\newblock \emph{arXiv: Quantum Physics}.

\bibitem[{Georgescu et~al.(2021)Georgescu, Barbalau, Ionescu, Khan, Popescu, and Shah}]{georgescu2021anomaly}
Georgescu, M.-I.; Barbalau, A.; Ionescu, R.~T.; Khan, F.~S.; Popescu, M.; and Shah, M. 2021.
\newblock Anomaly detection in video via self-supervised and multi-task learning.
\newblock In \emph{Proceedings of the IEEE/CVF conference on computer vision and pattern recognition}, 12742--12752.

\bibitem[{G{\"o}rnitz et~al.(2013)G{\"o}rnitz, Kloft, Rieck, and Brefeld}]{gornitz2013toward}
G{\"o}rnitz, N.; Kloft, M.; Rieck, K.; and Brefeld, U. 2013.
\newblock Toward supervised anomaly detection.
\newblock \emph{Journal of Artificial Intelligence Research}, 46: 235--262.

\bibitem[{Guo et~al.(2022)Guo, Liu, Li, Li, Gao, Qin, and Wen}]{guo2022quantum}
Guo, M.; Liu, H.; Li, Y.; Li, W.; Gao, F.; Qin, S.; and Wen, Q. 2022.
\newblock Quantum algorithms for anomaly detection using amplitude estimation.
\newblock \emph{Physica A: Statistical Mechanics and its Applications}, 604: 127936.

\bibitem[{Guo et~al.(2023)Guo, Pan, Li, Gao, Qin, Yu, Zhang, and Wen}]{guo2023quantum}
Guo, M.; Pan, S.; Li, W.; Gao, F.; Qin, S.; Yu, X.; Zhang, X.; and Wen, Q. 2023.
\newblock Quantum algorithm for unsupervised anomaly detection.
\newblock \emph{Physica A: Statistical Mechanics and its Applications}, 625: 129018.

\bibitem[{Han et~al.(2022)Han, Hu, Huang, Jiang, and Zhao}]{han2022adbench}
Han, S.; Hu, X.; Huang, H.; Jiang, M.; and Zhao, Y. 2022.
\newblock Adbench: Anomaly detection benchmark.
\newblock \emph{Advances in Neural Information Processing Systems}, 35: 32142--32159.

\bibitem[{Kim et~al.(2023)Kim, McCaskey, Heim, Modani, Stanwyck, and Costa}]{10247886}
Kim, J.-S.; McCaskey, A.; Heim, B.; Modani, M.; Stanwyck, S.; and Costa, T. 2023.
\newblock CUDA Quantum: The Platform for Integrated Quantum-Classical Computing.
\newblock In \emph{2023 60th ACM/IEEE Design Automation Conference (DAC)}, 1--4.

\bibitem[{Kong and Ramanan(2021)}]{kong2021opengan}
Kong, S.; and Ramanan, D. 2021.
\newblock Opengan: Open-set recognition via open data generation.
\newblock In \emph{Proceedings of the IEEE/CVF International Conference on Computer Vision}, 813--822.

\bibitem[{Kottmann et~al.(2021)Kottmann, Metz, Fraxanet, and Baldelli}]{kottmann2021variational}
Kottmann, K.; Metz, F.; Fraxanet, J.; and Baldelli, N. 2021.
\newblock Variational quantum anomaly detection: Unsupervised mapping of phase diagrams on a physical quantum computer.
\newblock \emph{Physical Review Research}, 3(4): 043184.

\bibitem[{Krizhevsky(2009)}]{cifar10}
Krizhevsky, A. 2009.
\newblock Learning Multiple Layers of Features from Tiny Images.

\bibitem[{LeCun et~al.(2010)LeCun, Cortes, Burges et~al.}]{lecun2010mnist}
LeCun, Y.; Cortes, C.; Burges, C.; et~al. 2010.
\newblock MNIST handwritten digit database.

\bibitem[{Li et~al.(2021{\natexlab{a}})Li, Sohn, Yoon, and Pfister}]{SAOE1}
Li, C.-L.; Sohn, K.; Yoon, J.; and Pfister, T. 2021{\natexlab{a}}.
\newblock Cutpaste: Self-supervised learning for anomaly detection and localization.
\newblock In \emph{Proceedings of the IEEE/CVF conference on computer vision and pattern recognition}, 9664--9674.

\bibitem[{Li et~al.(2021{\natexlab{b}})Li, Sohn, Yoon, and Pfister}]{li2021cutpaste}
Li, C.-L.; Sohn, K.; Yoon, J.; and Pfister, T. 2021{\natexlab{b}}.
\newblock Cutpaste: Self-supervised learning for anomaly detection and localization.
\newblock In \emph{Proceedings of the IEEE/CVF conference on computer vision and pattern recognition}, 9664--9674.

\bibitem[{Lin et~al.(2017)Lin, Goyal, Girshick, He, and Doll{\'a}r}]{Flos}
Lin, T.-Y.; Goyal, P.; Girshick, R.; He, K.; and Doll{\'a}r, P. 2017.
\newblock Focal loss for dense object detection.
\newblock In \emph{Proceedings of the IEEE international conference on computer vision}, 2980--2988.

\bibitem[{Liu and Rebentrost(2018)}]{liu2018quantum}
Liu, N.; and Rebentrost, P. 2018.
\newblock Quantum machine learning for quantum anomaly detection.
\newblock \emph{Physical Review A}, 97(4): 042315.

\bibitem[{Liu et~al.(2019)Liu, Luo, Li, Zhao, Gao et~al.}]{MLEP}
Liu, W.; Luo, W.; Li, Z.; Zhao, P.; Gao, S.; et~al. 2019.
\newblock Margin Learning Embedded Prediction for Video Anomaly Detection with A Few Anomalies.
\newblock In \emph{IJCAI}, volume~3, 023--3.

\bibitem[{Liznerski et~al.(2020)Liznerski, Ruff, Vandermeulen, Franks, Kloft, and M{\"u}ller}]{SAOE2}
Liznerski, P.; Ruff, L.; Vandermeulen, R.~A.; Franks, B.~J.; Kloft, M.; and M{\"u}ller, K.-R. 2020.
\newblock Explainable deep one-class classification.
\newblock \emph{arXiv preprint arXiv:2007.01760}.

\bibitem[{Moro and Prati(2023)}]{moro2023anomaly}
Moro, L.; and Prati, E. 2023.
\newblock Anomaly detection speed-up by quantum restricted Boltzmann machines.
\newblock \emph{Communications Physics}, 6(1): 269.

\bibitem[{Nielsen and Chuang(2010{\natexlab{a}})}]{nielsen2010quantum}
Nielsen, M.~A.; and Chuang, I.~L. 2010{\natexlab{a}}.
\newblock \emph{Quantum computation and quantum information}.
\newblock Cambridge university press.

\bibitem[{Nielsen and Chuang(2010{\natexlab{b}})}]{Nielsen_Chuang_2010}
Nielsen, M.~A.; and Chuang, I.~L. 2010{\natexlab{b}}.
\newblock \emph{Quantum Computation and Quantum Information: 10th Anniversary Edition}.
\newblock Cambridge University Press.

\bibitem[{Pan et~al.(2023)Pan, Zhu, Atici, and Cetin}]{pan2023hybrid}
Pan, H.; Zhu, X.; Atici, S.~F.; and Cetin, A. 2023.
\newblock A hybrid quantum-classical approach based on the hadamard transform for the convolutional layer.
\newblock In \emph{International Conference on Machine Learning}, 26891--26903. PMLR.

\bibitem[{Pang et~al.(2021)Pang, Ding, Shen, and Hengel}]{pang2021explainable}
Pang, G.; Ding, C.; Shen, C.; and Hengel, A. v.~d. 2021.
\newblock Explainable Deep Few-shot Anomaly Detection with Deviation Networks.
\newblock \emph{arXiv preprint arXiv:2108.00462}.

\bibitem[{Pang, Shen, and Van Den~Hengel(2019)}]{pang2019deep}
Pang, G.; Shen, C.; and Van Den~Hengel, A. 2019.
\newblock Deep anomaly detection with deviation networks.
\newblock In \emph{Proceedings of the 25th ACM SIGKDD international conference on knowledge discovery \& data mining}, 353--362.

\bibitem[{Peng et~al.(2024)Peng, Li, Liang, and Wang}]{peng2024hyq2}
Peng, Y.; Li, X.; Liang, Z.; and Wang, Y. 2024.
\newblock HyQ2: A Hybrid Quantum Neural Network for NextG Vulnerability Detection.
\newblock \emph{IEEE Transactions on Quantum Engineering}.

\bibitem[{Preskill(2018)}]{preskill2018quantum}
Preskill, J. 2018.
\newblock Quantum computing in the NISQ era and beyond.
\newblock \emph{Quantum}, 2: 79.

\bibitem[{Ruff et~al.(2019)Ruff, Vandermeulen, G{\"o}rnitz, Binder, M{\"u}ller, M{\"u}ller, and Kloft}]{ruff2019deep}
Ruff, L.; Vandermeulen, R.~A.; G{\"o}rnitz, N.; Binder, A.; M{\"u}ller, E.; M{\"u}ller, K.-R.; and Kloft, M. 2019.
\newblock Deep semi-supervised anomaly detection.
\newblock \emph{arXiv preprint arXiv:1906.02694}.

\bibitem[{Salehi et~al.(2021)Salehi, Sadjadi, Baselizadeh, Rohban, and Rabiee}]{salehi2021multiresolutionheadct}
Salehi, M.; Sadjadi, N.; Baselizadeh, S.; Rohban, M.~H.; and Rabiee, H.~R. 2021.
\newblock Multiresolution knowledge distillation for anomaly detection.
\newblock In \emph{Proceedings of the IEEE/CVF conference on computer vision and pattern recognition}, 14902--14912.

\bibitem[{Silvestre-Blanes et~al.(2019)Silvestre-Blanes, Albero-Albero, Miralles, P{\'e}rez-Llor{\'e}ns, and Moreno}]{SilvestreBlanes2019APFAITEX}
Silvestre-Blanes, J.; Albero-Albero, T.; Miralles, I.; P{\'e}rez-Llor{\'e}ns, R.; and Moreno, J. 2019.
\newblock A Public Fabric Database for Defect Detection Methods and Results.
\newblock \emph{Autex Research Journal}, 19: 363 -- 374.

\bibitem[{Sultani, Chen, and Shah(2018)}]{sultani2018real}
Sultani, W.; Chen, C.; and Shah, M. 2018.
\newblock Real-world anomaly detection in surveillance videos.
\newblock In \emph{Proceedings of the IEEE conference on computer vision and pattern recognition}, 6479--6488.

\bibitem[{Sun et~al.(2020)Sun, Yang, Zhang, Ling, and Peng}]{sun2020conditional}
Sun, X.; Yang, Z.; Zhang, C.; Ling, K.-V.; and Peng, G. 2020.
\newblock Conditional gaussian distribution learning for open set recognition.
\newblock In \emph{Proceedings of the IEEE/CVF Conference on Computer Vision and Pattern Recognition}, 13480--13489.

\bibitem[{Tabernik et~al.(2019)Tabernik, Šela, Skvarč, and Skočaj}]{Tabernik_2019SDD}
Tabernik, D.; Šela, S.; Skvarč, J.; and Skočaj, D. 2019.
\newblock Segmentation-based deep-learning approach for surface-defect detection.
\newblock \emph{Journal of Intelligent Manufacturing}, 31(3): 759–776.

\bibitem[{Tack et~al.(2020)Tack, Mo, Jeong, and Shin}]{SAOE3}
Tack, J.; Mo, S.; Jeong, J.; and Shin, J. 2020.
\newblock Csi: Novelty detection via contrastive learning on distributionally shifted instances.
\newblock \emph{Advances in neural information processing systems}, 33: 11839--11852.

\bibitem[{Tang et~al.(2024)Tang, Hua, Gao, Zhao, and Li}]{tang2024gadbench}
Tang, J.; Hua, F.; Gao, Z.; Zhao, P.; and Li, J. 2024.
\newblock Gadbench: Revisiting and benchmarking supervised graph anomaly detection.
\newblock \emph{Advances in Neural Information Processing Systems}, 36.

\bibitem[{Tian et~al.(2021)Tian, Pang, Chen, Singh, Verjans, and Carneiro}]{tian2021weakly}
Tian, Y.; Pang, G.; Chen, Y.; Singh, R.; Verjans, J.~W.; and Carneiro, G. 2021.
\newblock Weakly-supervised video anomaly detection with robust temporal feature magnitude learning.
\newblock In \emph{Proceedings of the IEEE/CVF international conference on computer vision}, 4975--4986.

\bibitem[{Wang et~al.(2023)Wang, Huang, Liu, Yi, Wu, and Wang}]{wang2023quantum}
Wang, M.; Huang, A.; Liu, Y.; Yi, X.; Wu, J.; and Wang, S. 2023.
\newblock A quantum-classical hybrid solution for deep anomaly detection.
\newblock \emph{Entropy}, 25(3): 427.

\bibitem[{Wei et~al.(2023)Wei, Liu, Xu, Shi, Shan, Zhao, and Gao}]{wei2023quantum}
Wei, L.; Liu, H.; Xu, J.; Shi, L.; Shan, Z.; Zhao, B.; and Gao, Y. 2023.
\newblock Quantum machine learning in medical image analysis: A survey.
\newblock \emph{Neurocomputing}, 525: 42--53.

\bibitem[{Wieler, Hahn, and Hamprecht(2023)}]{wieler_2023_8086136}
Wieler, M.; Hahn, T.; and Hamprecht, F.~A. 2023.
\newblock {Weakly Supervised Learning for Industrial Optical Inspection}.

\bibitem[{Yao et~al.(2023)Yao, Li, Zhang, Sun, and Zhang}]{Yao_2023_CVPR}
Yao, X.; Li, R.; Zhang, J.; Sun, J.; and Zhang, C. 2023.
\newblock Explicit Boundary Guided Semi-Push-Pull Contrastive Learning for Supervised Anomaly Detection.
\newblock In \emph{Proceedings of the IEEE/CVF Conference on Computer Vision and Pattern Recognition (CVPR)}, 24490--24499.

\bibitem[{Yoshihashi et~al.(2019)Yoshihashi, Shao, Kawakami, You, Iida, and Naemura}]{yoshihashi2019classification}
Yoshihashi, R.; Shao, W.; Kawakami, R.; You, S.; Iida, M.; and Naemura, T. 2019.
\newblock Classification-reconstruction learning for open-set recognition.
\newblock In \emph{Proceedings of the IEEE/CVF Conference on Computer Vision and Pattern Recognition}, 4016--4025.

\bibitem[{Zhang et~al.(2020)Zhang, Li, Guo, and Guo}]{zhang2020hybrid}
Zhang, H.; Li, A.; Guo, J.; and Guo, Y. 2020.
\newblock Hybrid models for open set recognition.
\newblock In \emph{Computer Vision--ECCV 2020: 16th European Conference, Glasgow, UK, August 23--28, 2020, Proceedings, Part III 16}, 102--117. Springer.

\bibitem[{Zhu et~al.(2024)Zhu, Ding, Tian, and Pang}]{zhu2024anomaly}
Zhu, J.; Ding, C.; Tian, Y.; and Pang, G. 2024.
\newblock Anomaly heterogeneity learning for open-set supervised anomaly detection.
\newblock In \emph{Proceedings of the IEEE/CVF Conference on Computer Vision and Pattern Recognition}, 17616--17626.

\end{thebibliography}

\clearpage

\begin{table*}[ht]
\centering
\caption{AUC results (mean ± std) on Mvtec AD datasets under the \textbf{general} setting (noted by $\heartsuit$). The best and second-best results and the third-best are respectively highlighted in \textcolor{red}{\textbf{red}} and \textcolor{blue}{\textbf{blue}} and \textbf{bold}. $\varpi$ is the number of anomaly classes.}

\resizebox{1\textwidth}{!}{
\begin{tabular}{l|c|c|c|c|c|c}

\toprule
\toprule
\textbf{Dataset} $(\varpi) $ & \textbf{SAOE}  & \textbf{FLOS} & \textbf{MLEP} & \textbf{DevNet} & \textbf{DRA} &  \textbf{DRA + Qsco} ($\ell =2 $) \\
\addlinespace[5pt]
\hline
\addlinespace[5pt]
\multicolumn{7}{c}{\textbf{Ten Anomaly Examples (Random)} $\heartsuit $ } \\
\hline
\addlinespace[5pt]
\textbf{Carpet} $(5)$ &  $0.755_{\pm 0.136}$ & $0.780_{\pm 0.009}$ & $0.781_{\pm 0.049}$ &  $\bm{0.850_{\pm 0.031}}$ & \textcolor{blue} { $\bm{0.908_{\pm 0.029}}$}  &  \textcolor{red} {$\bm{0.931_{\pm 0.024}}$}  \\

\addlinespace[5pt]
\textbf{Grid} $(5)$  &  $0.952_{\pm 0.011}$  &  $0.966_{\pm 0.005}$  &  $0.980_{\pm 0.009}$  &  $\bm{0.983_{\pm 0.009}}$  & \textcolor{blue} { $\bm{0.985_{\pm 0.014}}$}  & \textcolor{red} { $ \bm{0.995_{\pm 0.011}}$ }     \\ 

\addlinespace[5pt]

\textbf{Leather} $(5)$ &  \textcolor{red} {$\bm{1.000_{\pm 0.000}}$} &  $0.993_{\pm 0.004}$  &  $0.813_{\pm 0.158}$ &  $0.968_{\pm 0.006}$ &   \textcolor{red} {  $\bm{1.000_{\pm 0.000}}$ }     &  \textcolor{red} {  $\bm{1.000_{\pm 0.000}}$ }    \\

\addlinespace[5pt]

\textbf{Tile} $(5)$ &  $0.944_{\pm 0.013}$ &  $0.952_{\pm 0.010}$ &  $0.988_{\pm 0.009}$ &  $\bm{0.992_{\pm 0.004}}$ & \textcolor{blue} {  $\bm{0.999_{\pm 0.002}}$ }   & \textcolor{red} { $\bm{0.999_{\pm 0.001}}$}\\

\addlinespace[5pt]

\textbf{Wood} $(5)$ &  $0.976_{\pm 0.031}$ &  \textcolor{red} {$\bm{1.000_{\pm 0.000}}$} &  \textcolor{blue} {$\bm{0.999_{\pm 0.002}}$} &  $0.994_{\pm 0.004}$  &  $0.993_{\pm 0.005}$   &  $\bm{0.998_{\pm 0.003}}$   \\

\addlinespace[5pt]

\textbf{Bottle} $(3)$  &  \textcolor{blue} {$\bm{0.998_{\pm 0.003}}$} &  $0.995_{\pm 0.002}$ &  $0.981_{\pm 0.004}$  &  $\bm{0.998_{\pm 0.002}}$ & \textcolor{red} {$\bm{1.000_{\pm 0.000}}$}  &  \textcolor{red} {$\bm{1.000_{\pm 0.000}}$}   \\

\addlinespace[5pt]

\textbf{Capsule} $(5)$ &  $0.850_{\pm 0.054}$ &  $\bm{0.902_{\pm 0.017}}$  &  $0.818_{\pm 0.063}$ &  $0.855_{\pm 0.043}$     &  \textcolor{blue} {$\bm{0.963_{\pm 0.013}}$} &  
\textcolor{red} { $\bm{0.986_{\pm 0.011}}$ }  \\

\addlinespace[5pt]

\textbf{Pill} $(7)$ &  $0.872_{\pm 0.049}$  &  \textcolor{red} {$\bm{0.929_{\pm 0.012}}$} &  $0.845_{\pm 0.048}$ &  $0.850_{\pm 0.025}$ &  $\bm{0.871_{\pm 0.029}}$ & \textcolor{blue} {  $ \bm{0.889_{\pm 0.033}}$ }    \\

\addlinespace[5pt]

\textbf{Transistor}  $(4)$ &  $0.860_{\pm 0.053}$ &  $0.862_{\pm 0.037}$  &  \textcolor{red} {$\bm{0.927_{\pm 0.043}}$} &  $0.823_{\pm 0.034}$ &   $\bm{0.879_{\pm 0.037}}$    &  \textcolor{blue} {$ \bm{0.882_{\pm 0.031}}$}   \\

\addlinespace[5pt]

\textbf{Zipper}  $(7)$ &  $0.995_{\pm 0.004}$ &  $0.990_{\pm 0.008}$  &  $0.965_{\pm 0.002}$ &  \textcolor{red} {$\bm{0.997_{\pm 0.002}}$ }& \textcolor{red} {  $\bm{0.997_{\pm 0.002}}$ }   & \textcolor{red} { $ \bm{0.997_{\pm 0.002}}$}  \\

\addlinespace[5pt]

\textbf{Cable}  $(8)$ &  $0.862_{\pm 0.022}$ &  $0.890_{\pm 0.063}$  &  $0.857_{\pm 0.062}$ &  $\bm{0.943_{\pm 0.016}}$ & \textcolor{blue} {  $\bm{0.942_{\pm 0.009}}$ }   & \textcolor{red} { $ \bm{0.954_{\pm 0.012}}$}  \\

\addlinespace[5pt]

\textbf{Hazelnut}  $(4)$ &  \textcolor{red} {$\bm{1.000_{\pm 0.000}}$ }  &  \textcolor{red} {$\bm{1.000_{\pm 0.000}}$ }   &  \textcolor{red} {$\bm{1.000_{\pm 0.000}}$ }  &  \textcolor{red} {$\bm{1.000_{\pm 0.000}}$ }  &\textcolor{red} {$\bm{1.000_{\pm 0.000}}$ } &\textcolor{red} {$\bm{1.000_{\pm 0.000}}$}   \\

\addlinespace[5pt]

\textbf{MetalNut}  $(4)$ &  $0.976_{\pm 0.013}$ &  $0.984_{\pm 0.004}$  &  $0.974_{\pm 0.009}$ &  \textcolor{blue} {  $\bm{0.999_{\pm 0.002}}$ } & \textcolor{blue} {  $\bm{0.999_{\pm 0.002}}$ }   & \textcolor{red} { $ \bm{0.999_{\pm 0.001}}$ } \\

\addlinespace[5pt]

\textbf{Screw}  $(5)$ &  $\bm{0.975_{\pm 0.023}}$ &  $0.940_{\pm 0.017}$  &  $0.899_{\pm 0.039}$ &  $0.912_{\pm 0.059}$ & \textcolor{blue} {  $\bm{0.981_{\pm 0.002}}$ }   &  \textcolor{red} {$ \bm{0.983_{\pm 0.008}}$ } \\

\addlinespace[5pt]

\textbf{Toothbrush}  $(1)$ &  $0.865_{\pm 0.062}$ &  \textcolor{blue} {$\bm{0.900_{\pm 0.008}}$ } &  $0.783_{\pm 0.048}$ &  $0.852_{\pm 0.040}$ &   $\bm{0.879_{\pm 0.018}}$    &  \textcolor{red} {$ \bm{0.905_{\pm 0.013}}$ } \\

\addlinespace[5pt]

\multicolumn{7}{c}{\textbf{One Training Anomaly Example (Random)} $\heartsuit$ } \\ 

\hline

\addlinespace[5pt]

\textbf{Carpet} $(5)$ &  $0.766_{\pm 0.098}$ & $0.755_{\pm 0.026}$ & $0.701_{\pm 0.091}$ &  $\bm{0.803_{\pm 0.026}}$ &  \textcolor{red} {$\bm{0.887_{\pm 0.050}}$ } &  \textcolor{blue} {$\bm{0.877_{\pm 0.056}}$}    \\

\addlinespace[5pt]
\textbf{Grid} $(5)$ &  $\bm{0.921_{\pm 0.032}}$  &  $0.871_{\pm 0.076}$  &  $0.839_{\pm 0.028}$  &  $0.808_{\pm 0.075}$  &\textcolor{red} {  $\bm{0.970_{\pm 0.020}}$ } & \textcolor{blue} { $ \bm{0.951_{\pm 0.016}}$ }      \\ 

\addlinespace[5pt]

\textbf{Leather} $(5)$ &  $\bm{0.996_{\pm 0.007}}$ &  $0.791_{\pm 0.057}$  &  $0.781_{\pm 0.020}$ &  $0.881_{\pm 0.060}$ & \textcolor{blue} {  $\bm{0.997_{\pm 0.003}}$ }  &  \textcolor{red} {  $\bm{0.998_{\pm 0.002}}$ }      \\

\addlinespace[5pt]

\textbf{Tile} $(5)$ &  $\bm{0.935_{\pm 0.034}}$ &  $0.787_{\pm 0.038}$ &  $0.927_{\pm 0.036}$ &  $0.931_{\pm 0.029}$ & \textcolor{blue} {  $\bm{0.955_{\pm 0.019}}$ }   & \textcolor{red} { $\bm{0.986_{\pm 0.011}}$ }    \\

\addlinespace[5pt]

\textbf{Wood} $(5)$ &  $\bm{0.948_{\pm 0.009}}$ &  $0.927_{\pm 0.065}$ &  $0.660_{\pm 0.142}$ &  $0.720_{\pm 0.069}$  & \textcolor{blue} { $\bm{0.990_{\pm 0.008}}$ }  &   \textcolor{red} {$\bm{0.992_{\pm 0.013}}$}  \\

\addlinespace[5pt]

\textbf{Bottle} $(3)$&  $0.989_{\pm 0.019}$ &  $0.975_{\pm 0.023}$ &  $0.927_{\pm 0.090}$  &  $\bm{0.994_{\pm 0.006}}$ & \textcolor{blue} { $\bm{0.999_{\pm 0.001}}$ }  &  \textcolor{red} {$\bm{1.000_{\pm 0.000}}$}    \\

\addlinespace[5pt]

\textbf{Capsule} $(5)$ &  $0.611_{\pm 0.019}$ &  $0.666_{\pm 0.020}$  &  $0.558_{\pm 0.075}$ & $\bm{0.746_{\pm 0.007}}$  & \textcolor{blue} {  $\bm{0.831_{\pm 0.057}}$} &  
\textcolor{red} { $\bm{0.877_{\pm 0.054}}$ }    \\

\addlinespace[5pt]

\textbf{Pill} $(7)$ &  $0.652_{\pm 0.078}$  &  $\bm{0.745_{\pm 0.064}}$ &  $0.656_{\pm 0.061}$ &  $0.741_{\pm 0.025}$ & \textcolor{blue} { $\bm{0.800_{\pm 0.030}}$ }& \textcolor{red} {  $ \bm{0.836_{\pm 0.043}}$ }    \\

\addlinespace[5pt]

\textbf{Transistor}  $(4)$ &  $0.680_{\pm 0.182}$ &  $\bm{0.709_{\pm 0.041}}$  &  $0.695_{\pm 0.124}$ &  $0.694_{\pm 0.065}$ & \textcolor{blue} {  $\bm{0.724_{\pm 0.053}}$ }   &\textcolor{red} {   $ \bm{0.738_{\pm 0.071}}$ }  \\

\addlinespace[5pt]

\textbf{Zipper}  $(7)$ &  $\bm{0.970_{\pm 0.033}}$ &  $0.885_{\pm 0.033}$  &  $0.856_{\pm 0.086}$ &  $0.832_{\pm 0.045}$ & \textcolor{blue} {  $\bm{0.988_{\pm 0.007}}$ }   &   \textcolor{red} {$ \bm{0.991_{\pm 0.008}}$}   \\

\addlinespace[5pt]

\textbf{Cable}  $(8)$ &  $\bm{0.819_{\pm 0.060}}$ &  $0.790_{\pm 0.039}$  &  $0.688_{\pm 0.017}$ &  $0.704_{\pm 0.059}$ & \textcolor{red} {  $\bm{0.868_{\pm 0.021}}$ }   & \textcolor{blue} { $ \bm{0.848_{\pm 0.021}}$}  \\

\addlinespace[5pt]

\textbf{Hazelnut}  $(4)$ &  $0.961_{\pm 0.042}$ & \textcolor{blue} { $\bm{0.976_{\pm 0.021}}$  }&  $0.704_{\pm 0.090}$ & \textcolor{red} { $\bm{0.977_{\pm 0.071}}$ }&   $0.967_{\pm 0.049}$    &  $ \bm{0.972_{\pm 0.027}}$   \\

\addlinespace[5pt]

\textbf{MetalNut}  $(4)$ &  $0.922_{\pm 0.033}$ &  $\bm{0.930_{\pm 0.022}}$  &  $0.878_{\pm 0.038}$ &  $0.850_{\pm 0.051}$ & \textcolor{blue} {  $\bm{0.958_{\pm 0.059}}$ }   &  \textcolor{red} {$ \bm{0.971_{\pm 0.017}}$} \\

\addlinespace[5pt]

\textbf{Screw}  $(5)$ &  $0.653_{\pm 0.074}$ &  $0.337_{\pm 0.091}$  &  $0.675_{\pm 0.294}$ &  $\bm{0.727_{\pm 0.090}}$ & \textcolor{blue} {  $\bm{0.895_{\pm 0.079}}$ }   &  \textcolor{red} {$ \bm{0.898_{\pm 0.060}}$ }   \\

\addlinespace[5pt]

\textbf{Toothbrush}  $(1)$ &  $0.686_{\pm 0.110}$ &  $0.731_{\pm 0.028}$  &  $0.617_{\pm 0.058}$ &  \textcolor{red} {$\bm{0.778_{\pm 0.051}}$ }&   $\bm{0.735_{\pm 0.044}}$    &  \textcolor{blue} {$ \bm{0.774_{\pm 0.035}}$} \\

\addlinespace[5pt]
 
\bottomrule

\bottomrule

\end{tabular}
}
\label{AUCgeneralsubmvtecad}
\end{table*}

\begin{table*}[ht]
\centering
\caption{AUC results (mean ± std) on five real-world AD datasets and known subsets under the \textbf{hard} setting (noted by $\diamondsuit$). The best and second-best results and the third-best are respectively highlighted in \textcolor{red}{\textbf{red}} and \textcolor{blue}{\textbf{blue}} and \textbf{bold}.}

\resizebox{1\textwidth}{!}{
\begin{tabular}{c|c|c|c|c|c|c|c}

\toprule
\toprule
\textbf{Dataset}& \textbf{Known Subset} & \textbf{SAOE}  & \textbf{FLOS} & \textbf{MLEP} & \textbf{DevNet} & \textbf{DRA} & \textbf{DRA + Qsco} ($\ell =2 $) \\
\addlinespace[5pt]
\hline
\addlinespace[5pt]
\multicolumn{8}{c}{\textbf{Ten Training Anomaly Examples (Random)} $\diamondsuit$ } \\
\hline
\addlinespace[5pt]

& \textbf{broken end}& \textcolor{blue} { $\bm{0.712_{\pm 0.068}}$} & $0.585_{\pm 0.037}$ &\textcolor{red} { $\bm{0.732_{\pm 0.065}}$} &  $0.658_{\pm 0.111}$ & $0.644_{\pm 0.050}$  &  $\bm{0.662_{\pm 0.061}}$  \\
\addlinespace[5pt]

\multirow{5}{*}{\textbf{AITEX}} & \textbf{broken pick} & $\bm{0.629_{\pm 0.012}}$ & $0.548_{\pm 0.054}$ & $0.555_{\pm 0.027}$ &  $0.585_{\pm 0.028}$ & \textcolor{blue} { $\bm{0.645_{\pm 0.053}}$}  &  \textcolor{red} {$\bm{0.657_{\pm 0.018}}$}  \\

\addlinespace[5pt]
& \textbf{cut selvage}& \textcolor{blue} {$\bm{0.770_{\pm 0.014}}$} & $0.745_{\pm 0.035}$ & $0.682_{\pm 0.025}$ &  $0.709_{\pm 0.039}$ &  $\bm{0.757_{\pm 0.028}}$  &  \textcolor{red} {$\bm{0.775_{\pm 0.025}}$}  \\

\addlinespace[5pt]
& \textbf{fuzzyball} & \textcolor{red} {$\bm{0.842_{\pm 0.026}}$} & $0.550_{\pm 0.082}$ & $0.677_{\pm 0.223}$ &  $0.734_{\pm 0.039}$ & \textcolor{blue} { $\bm{0.805_{\pm 0.040}}$}  &  $\bm{0.763_{\pm 0.057}}$  \\
\addlinespace[5pt]

& \textbf{nep}& \textcolor{blue} { $\bm{0.771_{\pm 0.032}}$ }& $0.746_{\pm 0.060}$ & $0.740_{\pm 0.052}$ &  \textcolor{red} {$\bm{0.810_{\pm 0.042}}$ }& $0.767_{\pm 0.037}$  &  $\bm{0.769_{\pm 0.035}}$  \\
\addlinespace[5pt]

& \textbf{weft crack} & $\bm{0.618_{\pm 0.172}}$ &  \textcolor{red} {$\bm{0.636_{\pm 0.051}}$ }& $0.370_{\pm 0.037}$ &  $0.599_{\pm 0.137}$ & $0.584_{\pm 0.085}$  &  \textcolor{blue} {$\bm{0.629_{\pm 0.032}}$}  \\
\addlinespace[5pt]
\hline 

\addlinespace[5pt]
& \textbf{color}& $0.467_{\pm 0.067}$ & $0.760_{\pm 0.005}$ & $0.698_{\pm 0.025}$ &  $\bm{0.767_{\pm 0.015}}$ & \textcolor{blue} { $\bm{0.865_{\pm 0.074}}$}  &  \textcolor{red} {$\bm{0.879_{\pm 0.045}}$}  \\
\addlinespace[5pt]

\multirow{5}{*}{\textbf{Carpet}} & \textbf{cut}& $0.793_{\pm 0.175}$ & $0.688_{\pm 0.059}$ & $0.653_{\pm 0.120}$ &  $\bm{0.819_{\pm 0.037}}$ & \textcolor{blue} { $\bm{0.927_{\pm 0.032}}$}  &  \textcolor{red} {$\bm{0.943_{\pm 0.031}}$}  \\

\addlinespace[5pt]
& \textbf{hole}& $\bm{0.831_{\pm 0.125}}$ & $0.733_{\pm 0.014}$ & $0.674_{\pm 0.076}$ &  $0.814_{\pm 0.038}$ & \textcolor{red} { $\bm{0.916_{\pm 0.029}}$}  &  \textcolor{blue} {$\bm{0.901_{\pm 0.087}}$}  \\

\addlinespace[5pt]
& \textbf{metal}& $\bm{0.883_{\pm 0.043}}$ & $0.678_{\pm 0.083}$ & $0.764_{\pm 0.061}$ &  $0.863_{\pm 0.022}$ & \textcolor{blue} { $\bm{0.908_{\pm 0.054}}$}  &  \textcolor{red} {$\bm{0.931_{\pm 0.037}}$}  \\

\addlinespace[5pt]
&\textbf{thread} & $0.834_{\pm 0.297}$ & $0.946_{\pm 0.005}$ & $0.967_{\pm 0.006}$ &  $\bm{0.972_{\pm 0.009}}$ & \textcolor{red} { $\bm{0.985_{\pm 0.008}}$}  &  \textcolor{blue} {$\bm{0.983_{\pm 0.018}}$}  \\
\addlinespace[5pt]

\hline

\addlinespace[5pt]
& \textbf{bent}& $0.901_{\pm 0.023}$ & $0.827_{\pm 0.075}$ & $\bm{0.956_{\pm 0.013}}$ &  $0.904_{\pm 0.022}$ & \textcolor{red} { $\bm{0.987_{\pm 0.013}}$}  &  \textcolor{blue} {$\bm{0.983_{\pm 0.013}}$}  \\
\addlinespace[5pt]

\multirow{3}{*}{\textbf{Metal nut}} & \textbf{color}& $0.879_{\pm 0.018}$ & \textcolor{red} {$\bm{0.978_{\pm 0.008}}$} & $0.945_{\pm 0.039}$ &  \textcolor{blue} {$\bm{0.978_{\pm 0.016}}$} & $0.957_{\pm 0.018}$  &  $\bm{0.959_{\pm 0.031}}$  \\

\addlinespace[5pt]

&\textbf{flip} & $0.795_{\pm 0.062}$ & \textcolor{blue} {$\bm{0.942_{\pm 0.009}}$} & $0.805_{\pm 0.057}$ &  \textcolor{red} {$\bm{0.987_{\pm 0.004}}$} &  $0.908_{\pm 0.032}$  &  $\bm{0.933_{\pm 0.017}}$  \\
\addlinespace[5pt]

& \textbf{scratch}& $0.845_{\pm 0.041}$ & $0.943_{\pm 0.002}$ & $0.805_{\pm 0.153}$ &  \textcolor{red} {$\bm{0.991_{\pm 0.017}}$} &  $\bm{0.944_{\pm 0.021}}$  &  \textcolor{blue} {$\bm{0.955_{\pm 0.020}}$}  \\
\addlinespace[5pt]

\hline
\addlinespace[5pt]

\multirow{2}{*}{\textbf{ELPV}} & \textbf{mono}& $0.569_{\pm 0.035}$ & $0.629_{\pm 0.072}$ & \textcolor{red} {$\bm{0.756_{\pm 0.045}}$ }&  $0.599_{\pm 0.040}$ & \textcolor{blue} { $\bm{0.667_{\pm 0.056}}$}  &  $\bm{0.652_{\pm 0.050}}$  \\
\addlinespace[5pt]

& \textbf{poly}& $\bm{0.796_{\pm 0.084}}$ & $0.662_{\pm 0.042}$ & $0.734_{\pm 0.078}$ &  \textcolor{blue} { $\bm{0.804_{\pm 0.022}}$} & $0.742_{\pm 0.043}$  &  \textcolor{red} {$\bm{0.809_{\pm 0.034}}$}  \\
\addlinespace[5pt]

\hline

\addlinespace[5pt]

& \textbf{Barretts}& $0.698_{\pm 0.037}$ & $0.764_{\pm 0.066}$ & $0.540_{\pm 0.014}$ & \textcolor{red} { $\bm{0.834_{\pm 0.012}}$} & \textcolor{blue} { $\bm{0.808_{\pm 0.023}}$}  &  $\bm{0.787_{\pm 0.021}}$  \\
\addlinespace[5pt]

\multirow{4}{*}{\textbf{Hyper-Kvasir}} & \textbf{Barretts short seg} & $0.661_{\pm 0.034}$ & \textcolor{red} {$\bm{0.810_{\pm 0.034}}$ }& $0.480_{\pm 0.107}$ & \textcolor{blue} { $\bm{0.799_{\pm 0.036}}$} &  $0.716_{\pm 0.076}$  &  $\bm{0.743_{\pm 0.027}}$  \\
\addlinespace[5pt]

& \textbf{Esophagitis a} &\textcolor{blue} { $\bm{0.820_{\pm 0.034}}$ }& $\bm{0.815_{\pm 0.022}}$ & $0.646_{\pm 0.036}$ &  \textcolor{red} {$\bm{0.844_{\pm 0.014}}$} &  $0.811_{\pm 0.035}$  &  $0.802_{\pm 0.030}$  \\
\addlinespace[5pt]

& \textbf{Esophagitis b d}& $0.611_{\pm 0.017}$ &  \textcolor{blue} {$\bm{0.754_{\pm 0.073}}$ }& $0.621_{\pm 0.042}$ & \textcolor{red} { $\bm{0.810_{\pm 0.015}}$ }& $\bm{0.721_{\pm 0.052}}$  &  $0.720_{\pm 0.034}$  \\
\addlinespace[5pt]

\bottomrule

\bottomrule

\end{tabular}
}
\label{hardsetting10}
\end{table*}

\begin{table*}[ht]
\centering
\caption{AUC results (mean ± std) on five real-world AD datasets and known subsets under the \textbf{hard} setting (noted by $\diamondsuit$). The best and second-best results and the third-best are respectively highlighted in \textcolor{red}{\textbf{red}} and \textcolor{blue}{\textbf{blue}} and \textbf{bold}.}

\resizebox{1\textwidth}{!}{
\begin{tabular}{c|c|c|c|c|c|c|c}

\toprule
\toprule
\textbf{Dataset}& \textbf{Known Subset} & \textbf{SAOE}  & \textbf{FLOS} & \textbf{MLEP} & \textbf{DevNet} & \textbf{DRA} &  \textbf{DRA + Qsco} ($\ell =2 $) \\
\addlinespace[5pt]
\hline
\addlinespace[5pt]
\multicolumn{8}{c}{\textbf{One Training Anomaly Example (Random)} $\diamondsuit$ } \\
\hline
\addlinespace[5pt]

& \textbf{broken end}& \textcolor{red} {$\bm{0.778_{\pm 0.068}}$} & $0.645_{\pm 0.030}$ & $0.441_{\pm 0.111}$ &   \textcolor{blue} {$\bm{0.712_{\pm 0.069}}$} & $\bm{0.663_{\pm 0.063}}$  &  $0.652_{\pm 0.052}$  \\
\addlinespace[5pt]

\multirow{5}{*}{\textbf{AITEX}} & \textbf{broken pick} & $\bm{0.644_{\pm 0.039}}$ & $0.598_{\pm 0.023}$ & $0.476_{\pm 0.070}$ &  $0.552_{\pm 0.003}$ & \textcolor{blue} { $\bm{0.716_{\pm 0.036}}$}  &  \textcolor{red} {$\bm{0.730_{\pm 0.019}}$}  \\

\addlinespace[5pt]
& \textbf{cut selvage}&$0.681_{\pm 0.077}$ & $\bm{0.694_{\pm 0.036}}$ & $0.434_{\pm 0.149}$ &  $0.689_{\pm 0.016}$ & \textcolor{blue} { $\bm{0.731_{\pm 0.082}}$}  &  \textcolor{red} {$\bm{0.753_{\pm 0.039}}$}  \\
\addlinespace[5pt]
& \textbf{fuzzyball}& $\bm{0.650_{\pm 0.064}}$ & $0.525_{\pm 0.043}$ & $0.525_{\pm 0.157}$ &  $0.617_{\pm 0.075}$ & \textcolor{blue} { $\bm{0.661_{\pm 0.078}}$}  &  \textcolor{red} { $\bm{0.667_{\pm 0.049}}$}  \\

\addlinespace[5pt]
& \textbf{nep} & $\bm{0.710_{\pm 0.044}}$ & \textcolor{red} {$\bm{0.734_{\pm 0.038}}$} & $0.517_{\pm 0.059}$ &  \textcolor{blue} { $\bm{0.722_{\pm 0.023}}$ }& $0.683_{\pm 0.077}$  &  $0.652_{\pm 0.024}$  \\

\addlinespace[5pt]
& \textbf{weft crack} & $0.582_{\pm 0.108}$ & $0.546_{\pm 0.114}$ & $0.400_{\pm 0.029}$ &  $\bm{0.586_{\pm 0.134}}$ & \textcolor{red} { $\bm{0.611_{\pm 0.089}}$}  &  \textcolor{blue} {$\bm{0.602_{\pm 0.057}}$}  \\
\addlinespace[5pt]
\hline 

\addlinespace[5pt]
& \textbf{color}& $\bm{0.763_{\pm 0.100}}$ & $0.467_{\pm 0.278}$ & $0.547_{\pm 0.056}$ &  $0.716_{\pm 0.085}$ & \textcolor{red} { $\bm{0.828_{\pm 0.100}}$}  &  \textcolor{blue} {$\bm{0.817_{\pm 0.105}}$}  \\
\addlinespace[5pt]

\multirow{5}{*}{\textbf{Carpet}} & \textbf{cut}& $0.664_{\pm 0.165}$ & $\bm{0.685_{\pm 0.007}}$ & $0.658_{\pm 0.056}$ &  $0.666_{\pm 0.035}$ & \textcolor{red} { $\bm{0.871_{\pm 0.071}}$}  &  \textcolor{blue} {$\bm{0.853_{\pm 0.113}}$}  \\

\addlinespace[5pt]
& \textbf{hole}& $\bm{0.772_{\pm 0.071}}$ & $0.594_{\pm 0.142}$ & $0.653_{\pm 0.065}$ &  $0.721_{\pm 0.067}$ & \textcolor{blue} { $\bm{0.855_{\pm 0.063}}$}  &  \textcolor{red} {$\bm{0.859_{\pm 0.079}}$}  \\

\addlinespace[5pt]
&\textbf{metal} & $0.780_{\pm 0.172}$ & $0.701_{\pm 0.028}$ & $0.706_{\pm 0.047}$ &  $\bm{0.819_{\pm 0.032}}$ & \textcolor{blue} { $\bm{0.901_{\pm 0.054}}$}  &  \textcolor{red} {$\bm{0.910_{\pm 0.034}}$}  \\
\addlinespace[5pt]

& \textbf{thread}& $0.787_{\pm 0.204}$ &\textcolor{blue} { $\bm{0.941_{\pm 0.005}}$ }& $0.831_{\pm 0.117}$ &  $0.912_{\pm 0.044}$ &  $\bm{0.934_{\pm 0.043}}$  &  \textcolor{red} {$\bm{0.964_{\pm 0.027}}$}  \\
\addlinespace[5pt]

\hline

\addlinespace[5pt]
& \textbf{Bent}& $\bm{0.864_{\pm 0.032}}$ & $0.851_{\pm 0.046}$ & $0.743_{\pm 0.013}$ &  $0.797_{\pm 0.048}$ & \textcolor{red} { $\bm{0.957_{\pm 0.021}}$}  &  \textcolor{blue} {$\bm{0.917_{\pm 0.036}}$}  \\
\addlinespace[5pt]

\multirow{3}{*}{\textbf{Metal nut}} & \textbf{color}& $0.857_{\pm 0.037}$ & $0.821_{\pm 0.059}$ & $0.835_{\pm 0.075}$ &  $\bm{0.909_{\pm 0.023}}$ & \textcolor{blue} { $\bm{0.903_{\pm 0.053}}$}  &  \textcolor{red} {$\bm{0.916_{\pm 0.035}}$}  \\

\addlinespace[5pt]

&\textbf{Flip} & $0.751_{\pm 0.090}$ & $0.799_{\pm 0.058}$ & $\bm{0.813_{\pm 0.031}}$ &  $0.764_{\pm 0.014}$ & \textcolor{blue} { $\bm{0.916_{\pm 0.021}}$}  &  \textcolor{red} {$\bm{0.935_{\pm 0.030}}$}  \\
\addlinespace[5pt]

& \textbf{Scratch}& $0.792_{\pm 0.075}$ & $\bm{0.947_{\pm 0.027}}$ & $0.907_{\pm 0.085}$ &  \textcolor{red} {$\bm{0.952_{\pm 0.052}}$} & \textcolor{blue} { $\bm{0.950_{\pm 0.016}}$}  &  $0.943_{\pm 0.047}$  \\
\addlinespace[5pt]
\hline

\addlinespace[5pt]

\multirow{2}{*}{\textbf{ELPV}} & \textbf{mono}& $0.563_{\pm 0.102}$ & $0.717_{\pm 0.025}$ & \textcolor{red} {$\bm{0.649_{\pm 0.027}}$} &  $\bm{0.634_{\pm 0.087}}$ & $0.623_{\pm 0.047}$  &  \textcolor{blue} {$\bm{0.640_{\pm 0.034}}$}  \\
\addlinespace[5pt]
& \textbf{poly}& \textcolor{blue} {$\bm{0.665_{\pm 0.173}}$} & \textcolor{red} {$\bm{0.665_{\pm 0.021}}$} & $0.483_{\pm 0.247}$ &  $\bm{0.662_{\pm 0.050}}$ &  $0.630_{\pm 0.065}$  &  $0.592_{\pm 0.029}$  \\
\addlinespace[5pt]

\hline

\addlinespace[5pt]

& \textbf{Barretts}& $0.382_{\pm 0.117}$ & $\bm{0.703_{\pm 0.040}}$ & $0.438_{\pm 0.111}$ &   $0.672_{\pm 0.014}$ &\textcolor{red} { $\bm{0.740_{\pm 0.067}}$ } & \textcolor{blue} { $\bm{0.714_{\pm 0.037}}$ }\\
\addlinespace[5pt]

\multirow{4}{*}{\textbf{Hyper-Kvasir}} & \textbf{Barretts short seg} & $0.367_{\pm 0.050}$ & $0.538_{\pm 0.033}$ & $0.532_{\pm 0.075}$ &  $\bm{0.604_{\pm 0.048}}$ & \textcolor{blue} { $\bm{0.648_{\pm 0.047}}$}  &  \textcolor{red} {$\bm{0.660_{\pm 0.027}}$}  \\
\addlinespace[5pt]

& \textbf{Esophagitis a} & $0.518_{\pm 0.063}$ & $0.536_{\pm 0.040}$ & $0.491_{\pm 0.084}$ &  $\bm{0.569_{\pm 0.051}}$ &  \textcolor{blue} {$\bm{0.700_{\pm 0.085}}$ } &  \textcolor{red} {$\bm{0.737_{\pm 0.034}}$ } \\
\addlinespace[5pt]

& \textbf{Esophagitis b d}& $0.358_{\pm 0.039}$ & $0.505_{\pm 0.039}$ & $0.457_{\pm 0.086}$ &  $\bm{0.536_{\pm 0.033}}$ & \textcolor{blue} { $\bm{0.618_{\pm 0.061}}$}  &  \textcolor{red} {$\bm{0.635_{\pm 0.048}}$}  \\
\addlinespace[5pt]

\bottomrule

\bottomrule

\end{tabular}
}
\label{hardsetting1}
\end{table*}

\begin{figure*}[ht]

\begin{center}
\centerline{\includegraphics[width=1\textwidth]{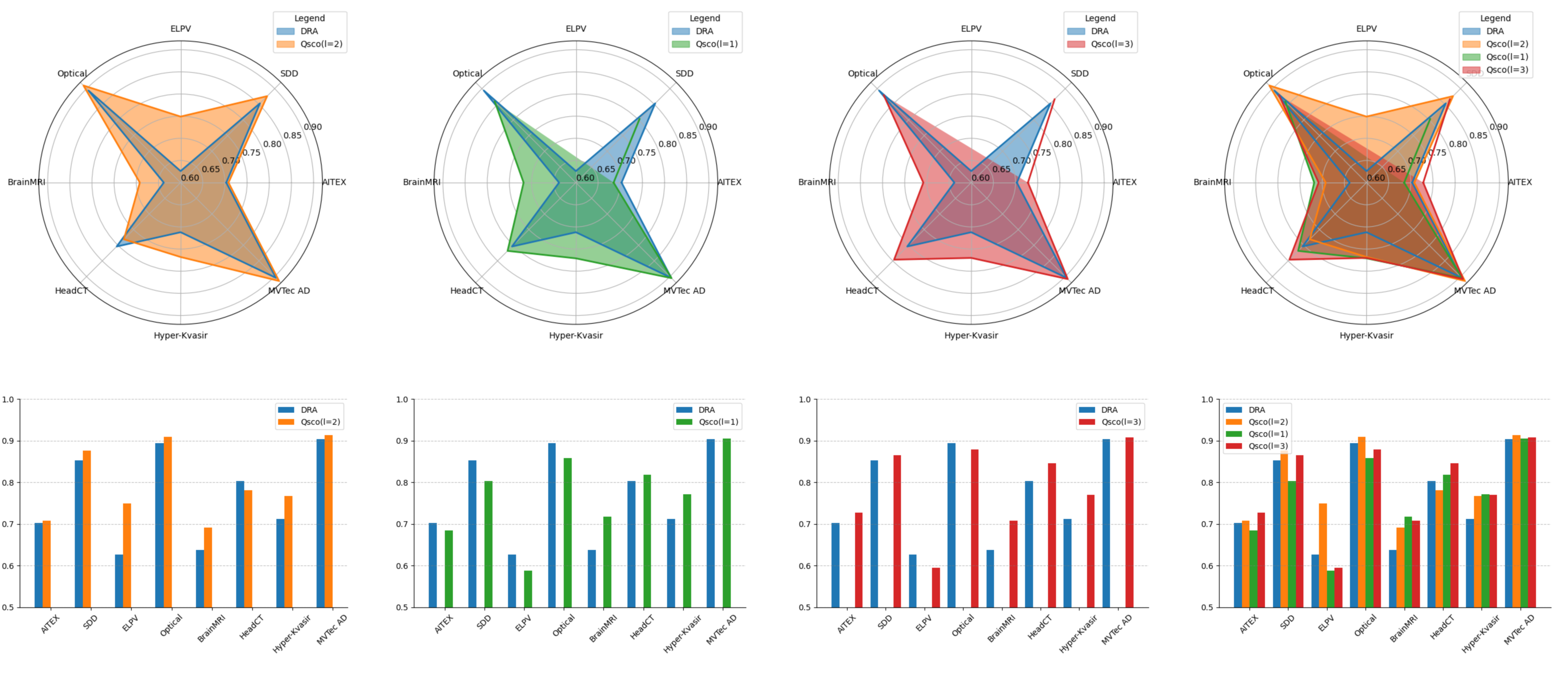}}
\caption{\textbf{Performance between DRA and Qsco + DRA with different $\ell$ for one training anomaly example}. The top row consists of radar charts, while the bottom row consists of corresponding bar charts, both illustrating the performance. The performance evaluation metric is AUC scores. From left to right, the charts correspond to $\ell = 2$,  $\ell = 1$, and $\ell = 3$. The last set of charts on the far right combines the results of all three $\ell$ values into a single visualization.}
\label{radar1}

\end{center}
\end{figure*}

\begin{figure*}[ht]

\begin{center}
\centerline{\includegraphics[width=1\textwidth]{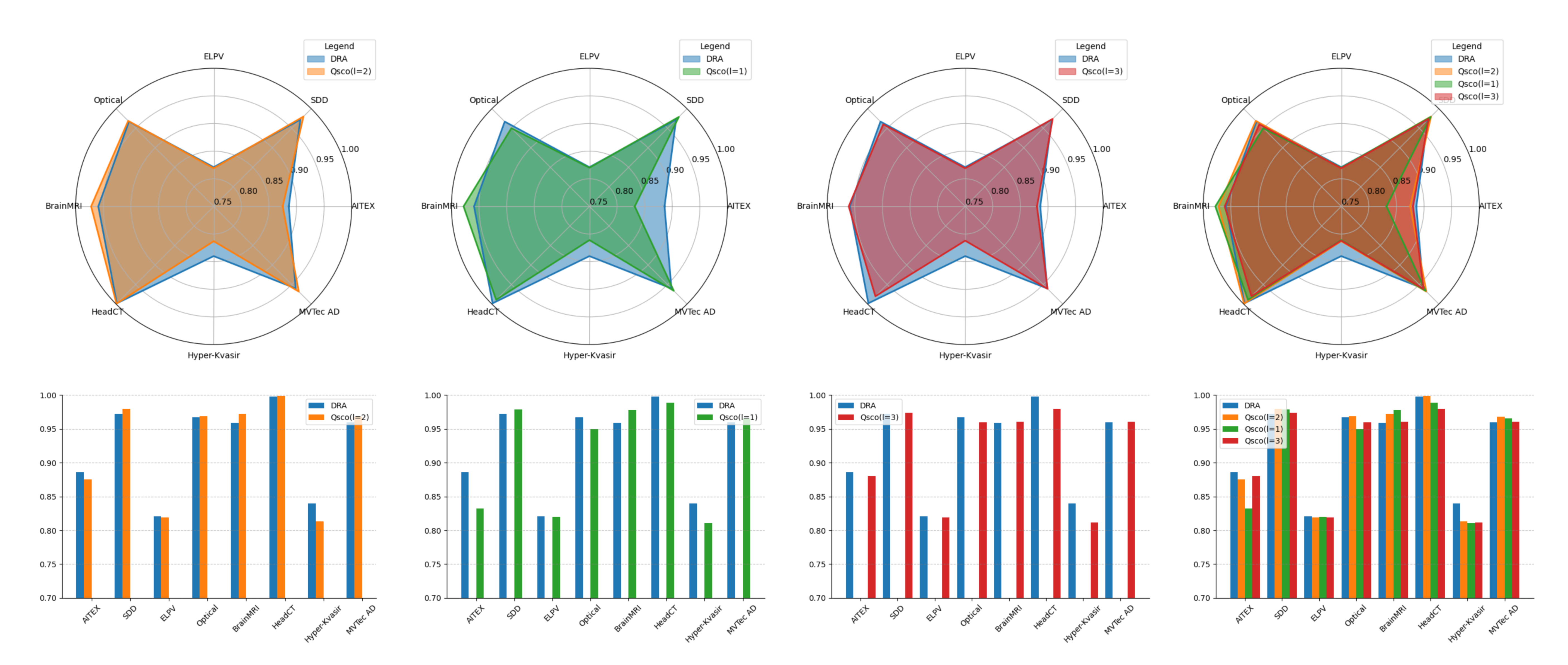}}
\caption{\textbf{Performance between DRA and Qsco + DRA with different $\ell$ for ten training anomaly examples}. The top row consists of radar charts, while the bottom row consists of corresponding bar charts, both illustrating the performance. The performance evaluation metric is AUC scores. From left to right, the charts correspond to $\ell = 2$,  $\ell = 1$, and $\ell = 3$. The last set of charts on the far right combines the results of all three $\ell$ values into a single visualization.}
\label{radar10}

\end{center}
\end{figure*}

\begin{table*}[h]
\centering
\caption{AUC results (mean ± std) on Mvtec AD datasets under the \textbf{general} setting (noted by $\heartsuit$). The increasing is highlighted from \colorbox{green!20}{low} to \colorbox{green!80}{high} and the decreasing (absolute values) is highlighted from \colorbox{red!20}{low} to \colorbox{red!80}{high} compared with $\ell =2 $. $\varpi$ is the number of anomaly classes. The amplitude damping noise model is $\clubsuit$, the phase flipping noise model is $\spadesuit $, and the depolarizing noise is noted as $\ast $, and the bit flipping noise is $\emptyset $. If the performance under the noise model is still better than the baseline (DRA), it is noted as $\flat$. SAOE, FLOS, and MLEP are from \citet{ding2022catching}.}

\resizebox{1\textwidth}{!}{
\begin{tabular}{l|c|c|c|c|c|c}

\toprule
\toprule
\textbf{Dataset} $(\varpi) $ &  \textbf{DRA} & \textbf{DRA + Qsco} ($\ell =2 $ )& \textbf{DRA + Qsco} ($\ell =2 $) $\clubsuit$ & \textbf{DRA + Qsco} ($\ell =2 $) $\spadesuit $& \textbf{DRA + Qsco} ($\ell =2 $) $\ast $& \textbf{DRA + Qsco} ($\ell =2 $) $\emptyset  $\\
\addlinespace[5pt]
\hline
\addlinespace[5pt]
\multicolumn{7}{c}{\textbf{Ten Anomaly Examples (Random)} $\heartsuit $ } \\
\hline
\addlinespace[5pt]
\textbf{Carpet} $(5)$ & $\bm{0.908_{\pm 0.029}}$  &  $\bm{0.931_{\pm 0.024}}$  & \colorbox{red!20}{$\bm{0.921_{\pm 0.041}}$  $ (1.07\%) \downarrow$ }$\flat $ & \colorbox{green!15}{ $\bm{0.934_{\pm 0.035}}$  $ (0.32\%) \uparrow$ }  $\flat$ 
 &\colorbox{red!21}{ $\bm{0.917_{\pm 0.042}}$  $ (1.50\%) \downarrow$ }$\flat$  &\colorbox{red!11}{ $\bm{0.927_{\pm 0.031}}$  $ (0.43\%) \downarrow$ }$\flat$\\

\addlinespace[5pt]
\textbf{Grid} $(5)$  &$\bm{0.985_{\pm 0.014}}$ & $ \bm{0.995_{\pm 0.011}}$  &\colorbox{red!22}{ $ \bm{0.982_{\pm 0.015}}$  $ (1.31\%) \downarrow$} & \colorbox{green!15}{ $\bm{0.998_{\pm 0.004}}$ $ (0.30\%) \uparrow$ }$\flat$ &  \colorbox{red!14}{$\bm{0.988_{\pm 0.010}}$ $ (0.70\%) \downarrow$ } $\flat$&  \colorbox{red!25}{$\bm{0.979_{\pm 0.010}}$ $ (1.61\%) \downarrow$ }  \\ 

\addlinespace[5pt]

\textbf{Leather} $(5)$ &   $\bm{1.000_{\pm 0.000}}$      &    $\bm{1.000_{\pm 0.000}}$  &   \colorbox{green!10}{  $\bm{1.000_{\pm 0.000}}$ $\longleftrightarrow$ } $\flat$ &  \colorbox{green!10}{  $\bm{1.000_{\pm 0.000}}$ $\longleftrightarrow$ }$\flat$ &  \colorbox{green!10}{  $\bm{1.000_{\pm 0.000}}$ $\longleftrightarrow$ }$\flat$ &  \colorbox{green!10}{  $\bm{1.000_{\pm 0.000}}$ $\longleftrightarrow$ }$\flat$  \\

\addlinespace[5pt]

\textbf{Tile} $(5)$ &    $\bm{0.999_{\pm 0.002}}$   &  $\bm{0.999_{\pm 0.001}}$  &  \colorbox{red!12}{ $\bm{0.991_{\pm 0.007}}$ $ (0.80\%) \downarrow$ } & \colorbox{red!11}{$\bm{0.993_{\pm 0.004}}$ $ (0.60\%) \downarrow$ } &\colorbox{red!16}{ $\bm{0.990_{\pm 0.006}}$ $ (0.90\%) \downarrow$} &\colorbox{red!17}{ $\bm{0.989_{\pm 0.008}}$ $ (1.00\%) \downarrow$}\\

\addlinespace[5pt]

\textbf{Wood} $(5)$ &  $\bm{0.993_{\pm 0.005}}$   &  $\bm{0.998_{\pm 0.003}}$ &  \colorbox{red!11}{ $\bm{0.995_{\pm 0.006}}$ $ (0.30\%) \downarrow$ } $\flat$&  \colorbox{red!10}{ $\bm{0.996_{\pm 0.005}}$ $ (0.20\%) \downarrow$ } $\flat$  &\colorbox{red!13}{   $\bm{0.993_{\pm 0.008}}$  $ (0.50\%) \downarrow$} $\flat$  & \colorbox{red!16}{  $\bm{0.989_{\pm 0.007}}$  $ (0.90\%) \downarrow$} \\

\addlinespace[5pt]

\textbf{Bottle} $(3)$  & $\bm{1.000_{\pm 0.000}}$  &  $\bm{1.000_{\pm 0.000}}$  &  \colorbox{green!10}{ $\bm{1.000_{\pm 0.000}}$ $\longleftrightarrow$} $\flat$ &  \colorbox{green!10}{ $\bm{1.000_{\pm 0.000}}$ $\longleftrightarrow$} $\flat$ &  \colorbox{green!10}{ $\bm{1.000_{\pm 0.000}}$ $\longleftrightarrow$} $\flat$ &  \colorbox{green!10}{ $\bm{1.000_{\pm 0.000}}$ $\longleftrightarrow$} $\flat$ \\

\addlinespace[5pt]

\textbf{Capsule} $(5)$ & $\bm{0.963_{\pm 0.013}}$ &  
 $\bm{0.986_{\pm 0.011}}$ &  
\colorbox{red!30}{ $\bm{0.964_{\pm 0.018}}$  $ (2.31\%) \downarrow$  } $\flat$&  
 \colorbox{red!22}{ $\bm{0.969_{\pm 0.017}}$  $ (1.72\%) \downarrow$  } $\flat$ &  
  \colorbox{red!42}{$\bm{0.941_{\pm 0.024}}$ $ (4.56\%) \downarrow$} &  
  \colorbox{red!40}{$\bm{0.942_{\pm 0.019}}$ $ (4.46\%) \downarrow$}\\

\addlinespace[5pt]

\textbf{Pill} $(7)$ &  $\bm{0.871_{\pm 0.029}}$ &  $ \bm{0.889_{\pm 0.033}}$    &  \colorbox{red!18}{ $ \bm{0.880_{\pm 0.014}}$   $ (1.01\%) \downarrow$ }$\flat$ &  \colorbox{green!40}{ $ \bm{0.938_{\pm 0.016}}$ $ (5.51\%) \uparrow$} $\flat$ &  \colorbox{green!30}{ $\bm{0.914_{\pm 0.027}}$ $ (2.81\%) \uparrow$ }$\flat$ &  \colorbox{green!21}{ $\bm{0.901_{\pm 0.012}}$ $ (1.35\%) \uparrow$ }$\flat$ \\

\addlinespace[5pt]

\textbf{Transistor}  $(4)$ &     $\bm{0.879_{\pm 0.037}}$    &  $ \bm{0.882_{\pm 0.031}}$ & \colorbox{green!50}{ $ \bm{0.949_{\pm 0.013}}$   $ (7.60\%) \uparrow$ } $\flat$& \colorbox{green!40}{  $\bm{0.932_{\pm 0.016}}$ $ (5.67\%) \uparrow$ } $\flat$  & \colorbox{green!50}{ $ \bm{0.950_{\pm 0.016}}$ $ (7.71\%) \uparrow$} $\flat$  & \colorbox{green!43}{ $ \bm{0.945_{\pm 0.014}}$ $ (7.14\%) \uparrow$} $\flat$ \\

\addlinespace[5pt]

\textbf{Zipper}  $(7)$ &   $\bm{0.997_{\pm 0.002}}$    &  $ \bm{0.999_{\pm 0.001}}$ &  \colorbox{green!10}{$ \bm{0.999_{\pm 0.001}}$ $\longleftrightarrow$} $\flat$&  \colorbox{green!10}{$ \bm{0.999_{\pm 0.001}}$ $\longleftrightarrow$} $\flat$ &   \colorbox{green!10}{$ \bm{0.999_{\pm 0.001}}$ $\longleftrightarrow$} $\flat$ &   \colorbox{green!10}{$ \bm{0.999_{\pm 0.001}}$ $\longleftrightarrow$} $\flat$\\

\addlinespace[5pt]

\textbf{Cable}  $(8)$ &   $\bm{0.942_{\pm 0.009}}$    &  $ \bm{0.954_{\pm 0.012}}$ & \colorbox{green!20}{   $ \bm{0.962_{\pm 0.012}}$ $ (0.84\%) \uparrow$ } $\flat$&  \colorbox{red!30}{ $ \bm{0.934_{\pm 0.018}}$ $ (2.10\%) \downarrow$ } &\colorbox{red!34}{ $\bm{0.930_{\pm 0.020}}$ $ (2.52\%) \downarrow$} &\colorbox{red!25}{ $\bm{0.938_{\pm 0.014}}$ $ (1.68\%) \downarrow$}\\

\addlinespace[5pt]

\textbf{Hazelnut}  $(4)$ & $\bm{1.000_{\pm 0.000}}$  & $\bm{1.000_{\pm 0.000}}$  & \colorbox{green!10}{ $\bm{1.000_{\pm 0.000}}$ $\longleftrightarrow$ } $\flat$ & \colorbox{green!10}{ $\bm{1.000_{\pm 0.000}}$ $\longleftrightarrow$ }  $\flat$ & \colorbox{green!10}{ $\bm{1.000_{\pm 0.000}}$ $\longleftrightarrow$ }  $\flat$ & \colorbox{green!10}{ $\bm{1.000_{\pm 0.000}}$ $\longleftrightarrow$ }  $\flat$  \\

\addlinespace[5pt]

\textbf{MetalNut}  $(4)$ &    $\bm{0.999_{\pm 0.002}}$    & $ \bm{0.999_{\pm 0.001}}$ & \colorbox{red!15}{  $ \bm{0.998_{\pm 0.002}}$ $ (0.10\%) \downarrow$ } & \colorbox{red!15}{  $ \bm{0.998_{\pm 0.003}}$ $ (0.10\%) \downarrow$ } & \colorbox{red!19}{  $ \bm{0.997_{\pm 0.003}}$ $ (0.20\%) \downarrow$ }  & \colorbox{red!19}{  $ \bm{0.997_{\pm 0.003}}$ $ (0.20\%) \downarrow$ }\\

\addlinespace[5pt]

\textbf{Screw}  $(5)$ &   $\bm{0.981_{\pm 0.002}}$    &  $ \bm{0.983_{\pm 0.008}}$ &  \colorbox{red!10}{  $ \bm{0.977_{\pm 0.016}}$ $ (0.61\%) \downarrow$ }
& \colorbox{green!13}{ $ \bm{0.984_{\pm 0.007}}$ $ (0.10\%) \uparrow$ } $\flat$ &  \colorbox{green!16}{ $\bm{0.987_{\pm 0.008}}$ $ (0.41\%) \uparrow$ } $\flat$ &  \colorbox{green!16}{ $\bm{0.987_{\pm 0.009}}$ $ (0.41\%) \uparrow$ } $\flat$ \\
\addlinespace[5pt]

\textbf{Toothbrush}  $(1)$ &    $\bm{0.879_{\pm 0.018}}$    &  $ \bm{0.905_{\pm 0.013}}$  & \colorbox{red!55}{  $ \bm{0.818_{\pm 0.020}}$ $ (9.61\%) \downarrow$ } &  \colorbox{red!45}{ $\bm{0.832_{\pm 0.024}}$ $ (8.10\%) \downarrow$ } &   \colorbox{red!40}{ $\bm{0.835_{\pm 0.042}}$ $ (7.73\%) \downarrow$ }&  \colorbox{red!47}{ $\bm{0.831_{\pm 0.024}}$ $ (8.18\%) \downarrow$ }\\

\addlinespace[5pt]

\multicolumn{7}{c}{\textbf{One Training Anomaly Example (Random)} $\heartsuit$ } \\ 

\hline

\addlinespace[5pt]

\textbf{Carpet} $(5)$  &  $\bm{0.887_{\pm 0.050}}$  &  $\bm{0.877_{\pm 0.056}}$  &  \colorbox{red!35}{ $\bm{0.838_{\pm 0.119}}$ $ (4.45\%) \downarrow$ } &\colorbox{green!44}{$\bm{0.918_{\pm 0.030}}$  $ (4.68\%) \uparrow$  } $\flat$ &  \colorbox{green!24}{$\bm{0.891_{\pm 0.066}}$  $ (1.60\%) \uparrow$} $\flat$ &  \colorbox{green!28}{$\bm{0.892_{\pm 0.058}}$  $ (1.71\%) \uparrow$} $\flat$ \\

\addlinespace[5pt]
\textbf{Grid} $(5)$ &  $\bm{0.970_{\pm 0.020}}$  &  $ \bm{0.951_{\pm 0.016}}$  & \colorbox{green!40}{ $ \bm{0.983_{\pm 0.012}}$  $ (3.36\%) \uparrow$  } $\flat$ & \colorbox{green!43}{ $\bm{0.989_{\pm 0.008}}$ $ (4.00\%) \uparrow$ } $\flat$ &  \colorbox{green!30}{$\bm{0.974_{\pm 0.018}}$ $ (2.42\%) \uparrow$ } $\flat$ &  \colorbox{red!13}{$\bm{0.949_{\pm 0.016}}$ $ (0.21\%) \uparrow$ } \\ 

\addlinespace[5pt]

\textbf{Leather} $(5)$ &    $\bm{0.997_{\pm 0.003}}$   &    $\bm{0.998_{\pm 0.002}}$ &   \colorbox{red!10}{ $\bm{0.998_{\pm 0.003}}$  $\longleftrightarrow$ } $\flat$ & \colorbox{red!15}{ $\bm{0.989_{\pm 0.008}}$  $ (0.90\%) \downarrow$ } &   \colorbox{red!12}{ $\bm{0.992_{\pm 0.005}}$ $ (0.60\%) \downarrow$ } & \colorbox{red!19}{ $\bm{0.985_{\pm 0.016}}$  $ (1.30\%) \downarrow$ } \\

\addlinespace[5pt]

\textbf{Tile} $(5)$ &  $\bm{0.955_{\pm 0.019}}$    & $\bm{0.986_{\pm 0.011}}$ & \colorbox{red!12}{ $\bm{0.981_{\pm 0.009}}$  $ (0.51\%) \downarrow$ }$\flat$ & \colorbox{green!14}{ $\bm{0.990_{\pm 0.008}}$ $ (0.41\%) \uparrow$ } $\flat$ &  \colorbox{green!10}{$\bm{0.986_{\pm 0.009}}$  $\longleftrightarrow$ } $\flat$  &  \colorbox{red!10}{$\bm{0.982_{\pm 0.010}}$ $ (0.41\%) \downarrow$ } $\flat$\\

\addlinespace[5pt]

\textbf{Wood} $(5)$ &  $\bm{0.990_{\pm 0.008}}$   &  $\bm{0.992_{\pm 0.013}}$ &  \colorbox{red!30}{ $\bm{0.960_{\pm 0.022}}$ $ (3.23\%) \downarrow$ }  &  \colorbox{red!12}{ $\bm{0.987_{\pm 0.014}}$ $ (0.50\%) \downarrow$ } & \colorbox{red!22}{ $\bm{0.969_{\pm 0.017}}$ $ (2.32\%) \downarrow$ } & \colorbox{red!32}{ $\bm{0.956_{\pm 0.020}}$ $ (3.63\%) \downarrow$ }\\ 

\addlinespace[5pt]

\textbf{Bottle} $(3)$&  $\bm{0.999_{\pm 0.001}}$   & $\bm{1.000_{\pm 0.000}}$   & \colorbox{green!10}{ $\bm{1.000_{\pm 0.000}}$ $\longleftrightarrow$ } $\flat$ & \colorbox{green!10}{ $\bm{1.000_{\pm 0.000}}$ $\longleftrightarrow$ } $\flat$ & \colorbox{green!10}{ $\bm{1.000_{\pm 0.000}}$ $\longleftrightarrow$ } $\flat$ & \colorbox{green!10}{ $\bm{1.000_{\pm 0.000}}$ $\longleftrightarrow$ } $\flat$\\

\addlinespace[5pt]

\textbf{Capsule} $(5)$ &    $\bm{0.831_{\pm 0.057}}$ &  
 $\bm{0.877_{\pm 0.054}}$  &  
 \colorbox{red!32}{ $\bm{0.853_{\pm 0.083}}$ $ (2.74\%) \downarrow$ } $\flat$ &  \colorbox{red!50}{ 
 $\bm{0.822_{\pm 0.027}}$$ (6.27\%) \downarrow$ } &  
  \colorbox{red!70}{$\bm{0.755_{\pm 0.083}}$ $ (13.91\%) \downarrow$} &  
  \colorbox{red!75}{$\bm{0.749_{\pm 0.059}}$ $ (14.60\%) \downarrow$}\\

\addlinespace[5pt]

\textbf{Pill} $(7)$ &   $\bm{0.800_{\pm 0.030}}$  &   $ \bm{0.836_{\pm 0.043}}$ &   \colorbox{red!38}{$ \bm{0.805_{\pm 0.034}}$  $ (3.71\%) \downarrow$ } $\flat$ &\colorbox{green!60}{   $ \bm{0.919_{\pm 0.033}}$ $ (9.93\%) \uparrow$ } $\flat$ &   \colorbox{green!39}{$\bm{0.886_{\pm 0.043}}$ $ (5.98\%) \uparrow$ } $\flat$ &   \colorbox{green!28}{$\bm{0.867_{\pm 0.044}}$ $ (3.71\%) \uparrow$ } $\flat$\\

\addlinespace[5pt]

 \textbf{Transistor}  $(4)$ &   $\bm{0.724_{\pm 0.053}}$    &   $ \bm{0.738_{\pm 0.071}}$  & \colorbox{green!43}{  $ \bm{0.786_{\pm 0.052}}$ $ (6.50\%) \uparrow$ } $\flat$ & \colorbox{green!48}{ $\bm{0.792_{\pm 0.032}}$ $ (7.32\%) \uparrow$ } $\flat$  &  \colorbox{green!43}{$ \bm{0.786_{\pm 0.039}}$ $ (6.50\%) \uparrow$ } $\flat$ &  \colorbox{green!15}{$ \bm{0.740_{\pm 0.055}}$ $ (0.27\%) \uparrow$ } $\flat$  \\

\addlinespace[5pt]

\textbf{Zipper}  $(7)$ &     $\bm{0.988_{\pm 0.007}}$    &   $ \bm{0.991_{\pm 0.008}}$   & \colorbox{red!11}{ $ \bm{0.988_{\pm 0.007}}$ $ (0.30\%) \downarrow$ } $\flat$ &  \colorbox{green!11}{   $\bm{0.993_{\pm 0.007}}$ $ (0.20\%) \uparrow$ } $\flat$ &  \colorbox{red!12}{ $ \bm{0.990_{\pm 0.009}}$ $ (0.10\%) \downarrow$ } &  \colorbox{red!17}{ $ \bm{0.982_{\pm 0.011}}$ $ (0.91\%) \downarrow$ } \\

\addlinespace[5pt]

\textbf{Cable}  $(8)$ &     $\bm{0.868_{\pm 0.021}}$    &  $ \bm{0.848_{\pm 0.021}}$  & \colorbox{green!40}{ $ \bm{0.876_{\pm 0.025}}$ $ (3.30\%) \uparrow$ } $\flat$ &   \colorbox{green!35}{  $\bm{0.867_{\pm 0.013}}$$ (2.24\%) \uparrow$ } &  \colorbox{red!15}{$\bm{0.840_{\pm 0.036}}$ $ (0.94\%) \downarrow$} &  \colorbox{red!30}{$\bm{0.819_{\pm 0.030}}$ $ (3.42\%) \downarrow$}  \\

\addlinespace[5pt]

\textbf{Hazelnut}  $(4)$ &   $\bm{0.967_{\pm 0.049}}$    &  $ \bm{0.972_{\pm 0.027}}$ &  \colorbox{green!22}{  $ \bm{0.985_{\pm 0.024}}$ $ (1.34\%) \uparrow$ } $\flat$ &  \colorbox{green!22}{  $ \bm{0.998_{\pm 0.006}}$ $ (1.34\%) \uparrow$ } $\flat$ & \colorbox{green!32}{  $\bm{0.999_{\pm 0.002}}$ $ (2.78\%) \uparrow$ } $\flat$ & \colorbox{green!32}{  $\bm{0.999_{\pm 0.001}}$ $ (2.78\%) \uparrow$ } $\flat$\\

\addlinespace[5pt]

\textbf{MetalNut}  $(4)$ &     $\bm{0.958_{\pm 0.059}}$    &  $ \bm{0.971_{\pm 0.017}}$  & \colorbox{red!36}{ $ \bm{0.933_{\pm 0.046}}$ $ (3.91\%) \downarrow$ } &  \colorbox{red!30}{    $\bm{0.938_{\pm 0.030}}$ $ (3.40\%) \downarrow$ } & \colorbox{red!28}{ $ \bm{0.936_{\pm 0.029}}$ $ (3.60\%) \downarrow$ } &  \colorbox{red!30}{    $\bm{0.939_{\pm 0.022}}$ $ (3.30\%) \downarrow$ }\\

\addlinespace[5pt]

\textbf{Screw}  $(5)$ &   $\bm{0.895_{\pm 0.079}}$   &  $ \bm{0.898_{\pm 0.060}}$   &  \colorbox{red!25}{ $ \bm{0.882_{\pm 0.068}}$ $ (1.78\%) \downarrow$  }&  \colorbox{red!30}{ $ \bm{0.880_{\pm 0.055}}$ $ (2.00\%) \downarrow$  } &  \colorbox{red!20}{ $\bm{0.888_{\pm 0.067}}$ $ (1.11\%) \downarrow$ } &  \colorbox{red!18}{ $\bm{0.889_{\pm 0.089}}$ $ (1.00\%) \downarrow$ }\\

\addlinespace[5pt]

\textbf{Toothbrush}  $(1)$ &    $\bm{0.735_{\pm 0.044}}$    &  $ \bm{0.774_{\pm 0.035}}$ & \colorbox{red!37}{ $ \bm{0.743_{\pm 0.041}}$ $ (4.01\%) \downarrow$ } $\flat$ & \colorbox{red!23}{ $ \bm{0.765_{\pm 0.035}}$ $ (1.16\%) \downarrow$ } $\flat$ & \colorbox{red!23}{ $ \bm{0.765_{\pm 0.037}}$ $ (1.16\%) \downarrow$ } $\flat$ & \colorbox{green!10}{ $\bm{0.774_{\pm 0.034}}$ $\longleftrightarrow$ } $\flat$ \\

\addlinespace[5pt]
 
\bottomrule

\bottomrule

\end{tabular}
}

\label{AUCgeneralsubmvtecadnoise}
\end{table*}

\end{document}